\documentclass[10pt,journal]{IEEEtran}
\ifCLASSOPTIONcompsoc
  \usepackage[nocompress]{cite}
\else
  \usepackage{cite}
\fi
\usepackage{amsmath,amsfonts}
\usepackage{algorithmic}
\usepackage{algorithm}
\usepackage{array}
\usepackage{caption}
\usepackage[caption=false,font=normalsize,labelfont=sf,textfont=sf]{subfig}
\usepackage{textcomp}
\usepackage{stfloats}
\usepackage{url}
\usepackage{verbatim}

\usepackage{graphicx}
\usepackage{boldline,multirow}
\usepackage{booktabs}
\usepackage{diagbox}
\usepackage{xcolor}

\usepackage{soul}
\usepackage{rotating}
\usepackage{enumitem}
\usepackage{pdfpages}
\usepackage{flushend}
\usepackage{hhline}
\usepackage[english]{babel}
\usepackage[square,sort,comma,numbers]{natbib}

\begin{document}
\newcommand{\lsx}[1]{{\bf\color{cyan}[{\sc Lsx:} #1]}}
\newcommand{\wyf}[1]{{\bf\color{red}[{\sc Wyf:} #1]}}

\newcommand{\modelcla}{MDCPM}
\newcommand{\modelsim}{AutoPT-Sim}

\title{A Unified Modeling Framework for Automated Penetration Testing}

\author{Yunfei Wang*, Shixuan Liu*, Wenhao Wang*, Changling Zhou, Chao Zhang,~\IEEEmembership{Member,~IEEE}, \\ Jiandong Jin, Cheng Zhu

\IEEEcompsocitemizethanks{
\IEEEcompsocthanksitem Yunfei Wang and Cheng Zhu are with the National Key Laboratory of Information Systems Engineering, National University of Defense Technology, Hunan, China. E-mail: \{wangyunfei, zhucheng\}@nudt.edu.cn
\IEEEcompsocthanksitem Shixuan Liu is with the Department of Intelligent Data Science, College of Computer Science and Technology, National University of Defense Technology, Hunan, China. E-mail: liushixuan@nudt.edu.cn
\IEEEcompsocthanksitem Wenhao Wang is with the National University of Defense Technology, Hunan, China. E-mail: wangwenhao11@nudt.edu.cn
\IEEEcompsocthanksitem Chao Zhang is with the Tsinghua University, Beijing, China. E-mail: 
chaoz@tsinghua.edu.cn
\IEEEcompsocthanksitem Jiandong Jin and Changling Zhou are with the Computer Center, Peking University, Beijing, China. E-mail: \{jiandong.jin, zclfly\}@pku.edu.cn
}

\thanks{*These authors contributed equally.}
\thanks{(Corresponding author: Cheng Zhu.)}
}


\markboth{IEEE Transactions on Information Forensics and Security}%
{Wang \MakeLowercase{\textit{et al.}}: A Unified Modeling Framework for Automated Penetration Testing}


\maketitle
\begin{abstract}
The integration of artificial intelligence into automated penetration testing (AutoPT) has highlighted the necessity of simulation modeling for the training of intelligent agents, due to its cost-efficiency and swift feedback capabilities. Despite the proliferation of AutoPT research, there is a recognized gap in the availability of a unified framework for simulation modeling methods. This paper presents a systematic review and synthesis of existing techniques, introducing \modelcla~to categorize studies based on literature objectives, network simulation complexity, dependency of technical and tactical operations, and scenario feedback and variation. To bridge the gap in unified method for multi-dimensional and multi-level simulation modeling, dynamic environment modeling, and the scarcity of public datasets, we introduce \modelsim, a novel modeling framework that based on policy automation and encompasses the combination of all sub dimensions. 
\modelsim~offers a comprehensive approach to modeling network environments, attackers, and defenders, transcending the constraints of static modeling and accommodating networks of diverse scales. We publicly release a generated standard network environment dataset and the code of Network Generator. By integrating publicly available datasets flexibly, support is offered for various simulation modeling levels focused on policy automation in \modelcla~and the network generator help researchers output customized target network data by adjusting parameters or fine-tuning the network generator.

\end{abstract}

\section{Introduction}
\IEEEPARstart{T}{he} Internet is crucial for modern life and social advancement, significantly impacting government, finance, energy, and military sectors, yet cybersecurity remains a critical concern despite enhanced convenience and services~\cite{chen2022research, wani2021sdn, verma2024revisiting}.
Penetration testing, which simulates hacker attacks to assess vulnerabilities and overall security~\cite{abu2018automated, dorchuck2021goal}, is vital for improving cybersecurity, but is complex and time consuming, relying on the expertise of the tester~\cite{applebaum2016intelligent}. As a solution, automated penetration testing (AutoPT) has emerged to replace human efforts and accelerate evaluations~\cite{ghanem2018reinforcement}.

AutoPT involves two main processes: intelligent decision-making and automatic execution. The intelligent decision-making phase is critical, encompassing target identification, attack path selection, and method determination, effectively replacing manual judgments. Current approaches to intelligent decision-making can be classified into three categories:

\begin{enumerate}
    \item \textit{Fixed Script Execution Methods} employs predetermined rules, which limits adaptability and flexibility when working with other penetration tools~\cite{hacks2021towards}.
    \item \textit{PDDL with Planner Methods} leverage the PDDL language to define penetration actions, with classical planners facilitating the process~\cite{applebaum2017analysis,yichao2019improved}. While PDDL enables detailed definition of penetration parameters and easy integration with tools, manually crafting these definitions is labor-intensive, requiring diverse approaches for various tools and tactics.
    \item \textit{Artificial Intelligence Methods} adopt attack trees~\cite{dorchuck2021goal}, attack graphs~\cite{obes2013attack}, reinforcement learning~\cite{hu2020automated,chowdhary2020autonomous}, and large language models~\cite{bianou2024pentest,deng2024pentestgpt} to guide decision-making in penetration testing, representing a leading trend in current research.
\end{enumerate}

For all of the above categories, it is important to model target scenarios, technical and tactical elements, and decision parameters. AI methods, in particular, require extensive training in diverse scenarios to reveal hidden patterns and improve generalizability. 

On the other hand, the automatic execution phase utilizes various penetration tools to implement predetermined actions, automating tasks typically performed by human experts. 
Integrating intelligent decision-making with automatic execution seeks to help experts in network penetration, improve efficiency, reduce costs, and improve security outcomes~\cite{chenke2023survey}.

\begin{figure*}[tb]
    \centering
    \includegraphics[width=\linewidth]{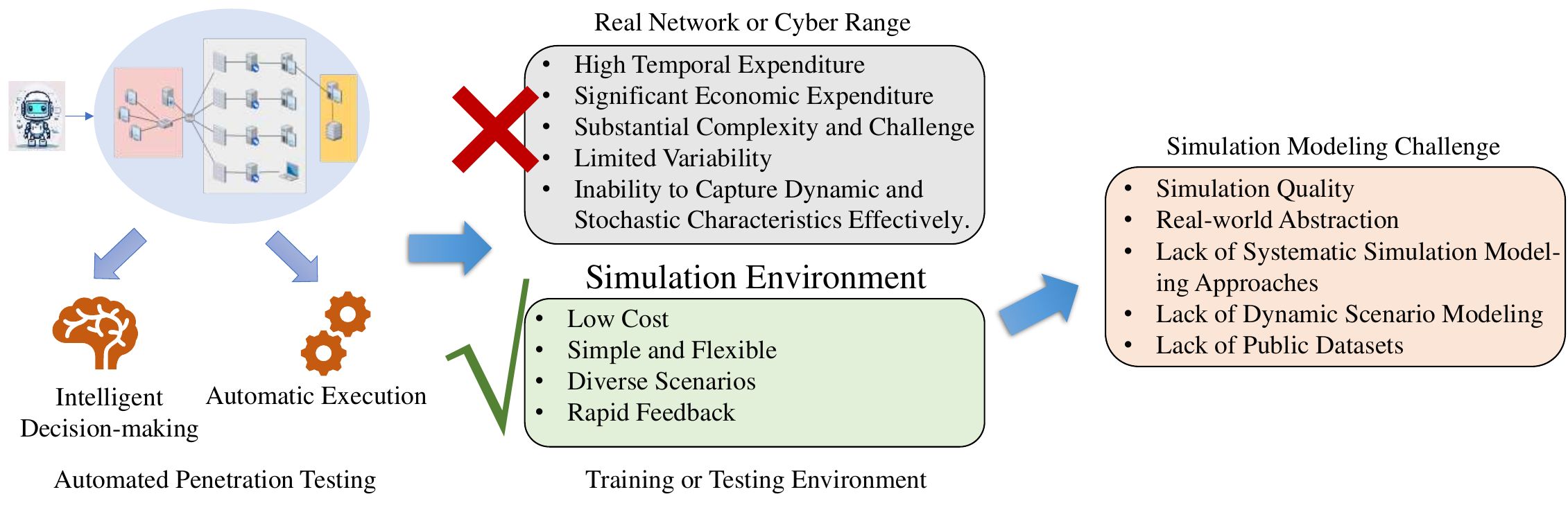}
    \caption{The Necessity and Challenges of Simulation Environments Modeling in AutoPT}
    \label{Introduction}
\end{figure*}

As shown in Figure~\ref{Introduction}, training within real network environments or cyber ranges demands significant resources, making it impractical due to high time and financial costs, as well as the complexities involved.
The inherent unpredictability and lack of reproducibility of real-world scenarios pose significant challenges for agents to in conducting repetitive training and recognizing patterns. By contrast, simulation environments provide a cost-effective and straightforward solution, with flexibility that allows for diverse scenarios and rapid feedback—key to improving the AutoPT decision-making process.

The quality and fidelity of simulation modeling is crucial for effective algorithm training in AutoPT. Real-world application demands algorithms that are both capable of learning and refining their performance from data and reflective of essential real-world features. 
Despite advances in AutoPT research, there is a lack of systematic analysis of simulation modeling methodologies. 
Current methods are diverse and fragmented, lacking a unified framework for modeling characteristics, elements, granularity, and procedures.  This hampers the development of efficient simulation models. 
A public dataset is essential for a rigorous and unbiased evaluation of penetration testing algorithms. Yet, concerns over security and privacy often result in a scarcity of public datasets for simulated networks, hindering a comprehensive assessment of AutoPT's efficacy.


Creating a unified framework presents several challenges. Firstly, the complexity of real-world systems demands models capable of encompassing diverse behaviors and interactions. Secondly, rapid technological advancements and evolving cyber threats require models that can be quickly updated. Thirdly, diverse research goals and stakeholder needs in AutoPT complicate the creation of a universally applicable framework. Lastly, privacy and security concerns must be prioritized, ensuring that models and datasets adhere to ethical standards while accurately depicting real-world attack scenarios and ensuring availability and reproducibility.


This paper comprehensively reviews academic literature on AutoPT Modeling from the 1990s to today, focusing on key terms including AutoPT, network attack-defense games, automated red teams, and related areas. We offer an in-depth analysis and interpretation of simulation modeling methods used in various AutoPT studies. Our contributions include:

\begin{itemize}
    \item We are the first to analyze and classify simulation modeling methods in AutoPT. Through an extensive review of the literature on AutoPT, we decompose the elements in AutoPT's simulation modeling and propose the \textbf{M}ulti-\textbf{D}imensional \textbf{C}lassification System for \textbf{P}enetration Testing \textbf{M}odeling (\modelcla) to systematically organize current simulation modeling methods within AutoPT.
    
    \item To address the absence of a unified approach for multi-dimensional and multi-level simulation modeling, dynamic environment modeling, and the lack of public datasets, we introduce the AutoPT Simulation Modeling Framework (\modelsim), a novel framework that leverages policy automation and integrates all sub-dimensions across the other three dimensions.

    \item We present a publicly available network simulation dataset along with the Network Generator code. This dataset can be flexibly combined to support various simulation modeling levels focused on policy automation within \modelcla. The Network Generator enables customizable network data generation by adjusting parameters or fine-tuning the settings, facilitating future research in AutoPT.
\end{itemize}

The paper is organized as follows: Section 2 examines the theoretical foundations and core modeling elements in simulation modeling for AutoPT. Section 3 presents our proposed Multi-Dimensional Penetration Testing Modeling Classification System (\modelcla), discussing typical cases and current research trends. Section 4 introduces AutoPT Simulation Modeling Framework (\modelsim), along with the open-source network simulation dataset and Network Generator. Section 5 concludes with a summary of our contributions and future work.

\section{THEORETICAL FOUNDATIONS OF AUTOPT SIMULATION MODELING}
\subsection{Control Theory Perspective}

In AutoPT, simulation modeling not only involves the abstraction and modeling of the target cyberspace but also encompasses the modeling of interactive entities within the network—attackers and defenders. We draw upon nonlinear system theory from Modern Control Theory to understand and categorize simulation modeling in AutoPT.



Modern control theory represents a significant domain within automatic control, dedicated to the analysis and design of complex systems, encompassing multiple-input-multiple-output, nonlinear, and time-varying configurations. This theoretical framework utilizes state variables and state space equations to depict the dynamic behavior of systems, thereby elucidating their internal states. Such an approach conceptualizes a system as a unified entity, highlighting the interdependence of its components, feedback mechanisms, hierarchical structures, and open systems characteristics. In the context of AutoPT, this perspective facilitates a thorough understanding and simulation of cybersecurity dynamics by conceptualizing the target network, attackers, and defenders as an integrated system. Control theory quantifies the status of the system through assignment of a state and formally describe the progression of AutoPT. It adopts a state-based approach to define cost structures that balance security and availability. The \textit{information state} in control theory compresses attacker-defender information to a level sufficient for optimal decision-making, converting determining the optimal defense policy into a sequential optimization problem~\cite{ErikMiehling2019Control}.


The complexity and variability of network activities, coupled with the multifactorial nature of penetration and defense strategies, means that identical actions on the same system can yield divergent outcomes, indicating that the penetration testing system is inherently dynamic. It implies that the same action combinations may produce different outputs, rendering linear differential equations inadequate for describing the system's behavior. The superposition principle is inapplicable, thus classifying such systems as a dynamic nonlinear system. Traditional research often models penetration testing as a discrete system based on the actions of attackers and defenders, treating these actions as discrete events. The system's output is influenced not only by its structural parameters but also by the inputs and initial conditions.



\subsection{AutoPT as a Dynamic Nonlinear System}
\subsubsection{System Components}

The modeling framework in AutoPT incorporates both the abstract representation of network environments and the simulation of adversarial interactions between attackers and defenders. Environmental modeling directly shapes the objectives, action spaces, and strategic decision-making processes of both agents. Concurrently, their adversarial interactions induce environmental modifications, establishing a dynamic feedback loop that subsequently shapes their adaptive strategic responses.

\begin{itemize}
    \item Network environment: The target network environment, encompassing its architecture and assets, serves as the foundational framework for attack-defense interactions, shaping the dynamics, realism, and reliability of simulated engagements. Accurate modeling of network architecture and assets is critical for achieving realistic simulations, as it captures the fluidity and complexity of real-world network conditions. This precision enables a more robust assessment of vulnerabilities and the efficacy of defensive measures.
    \item Attackers and defenders: AutoPT simulates adversarial tactics, techniques, and procedures to identify security vulnerabilities in target networks by modeling attacker and defender behaviors~\cite{chenke2023survey}. The attacker model requires explicit specification of objectives and permissible actions for simulated agents, while defender modeling accommodates greater flexibility, incorporating both static strategies (e.g., firewall policies, intrusion-detection system configurations~\cite{furfaro2017using}) and dynamic defense mechanisms~\cite{applebaum2016intelligent}. Observations for both parties are characterized by incomplete information and noise, including missed detections and false alarms~\cite{ErikMiehling2019Control}. We assume an information structure adhering to the \textit{perfect recall} assumption, wherein attackers and defenders retain complete historical records of past observations and decisions. Consequently, defenders at time $t$ maintain access to all prior historical data for decision-making.
\end{itemize}
In partially observable environments involving attackers and defenders, two primary approaches exist for modeling their interactions~\cite{ErikMiehling2019Control}: probabilistic uncertainty and nondeterministic uncertainty.
\subsubsection{System Dynamics}

In the AutoPT system, randomness and dynamic behavior arise from three interdependent factors: adversarial interactions between attackers and defenders, fluctuations in network environments, and stochastic events. System dynamism is thus inherently tied to these interdependent factors.

\begin{itemize}
    \item Input-Output. Attacker and defender actions exhibit bidirectional coupling: their interdependent decisions act as both system inputs and state-dependent outputs, driving iterative state evolution.
    \item State Transitions. The state \( S(t+1)\) evolves stochastically based on \( S(t) \), concurrent actions \( A^A(t) \) (attacker) and \( A^D(t) \) (defender), and exogenous stochastic events $W(t)$.
    \item Feedback Mechanisms. The utility function \( U(t) = (U^A(t), U^D(t)) \) critically shapes system dynamics by providing post-action feedback to both agents. This function serves dual roles: predefined reward signals or emergent behavioral outputs.
\end{itemize}

\subsubsection{Formal expression}

\begin{figure*}[tb]
    \centering
    \includegraphics[width=150mm]{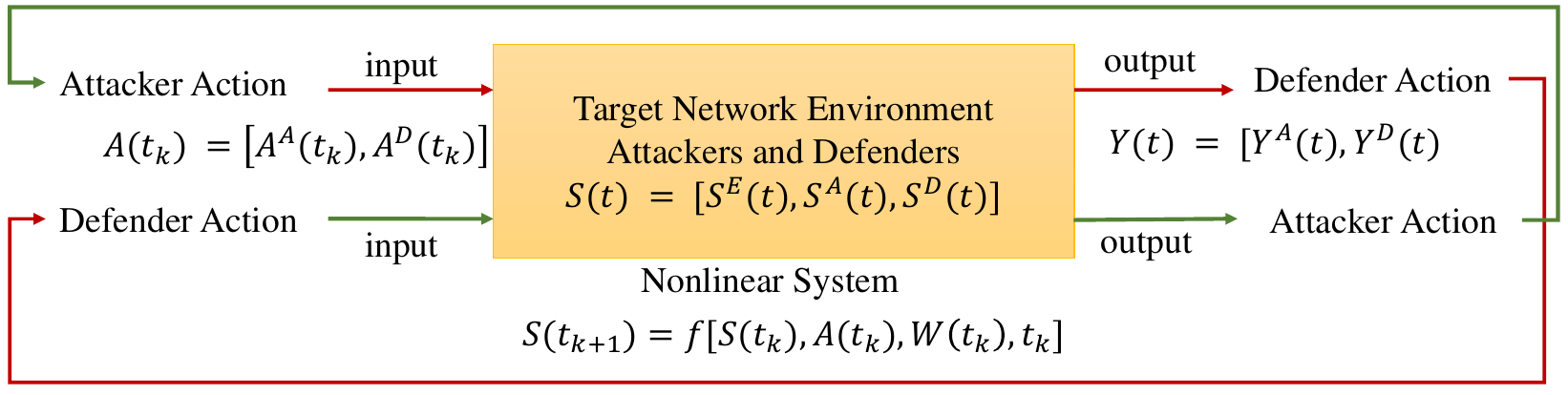}
    \caption{The Dynamic Nonlinear System Framework of AutoPT Simulation Modeling}
    \label{nonlinear}
\end{figure*}

%

As illustrated in Figure~\ref{nonlinear}, the AutoPT system comprises a target network environment, adversarial agents (attackers/defenders), and a state vector $S(t) = [S^E(t), S^A(t), S^D(t)]$, where $S^E(t)$, $S^A(t)$ and $S^D(t)$ denote the network, attacker, and defender states, respectively.
Stochastic disturbances $W(t)$, representing exogenous factors (e.g., hardware failures, user errors, or action success probabilities), introduce uncertainty into state transitions. These disturbances induce non-deterministic critical outputs even under identical inputs. For $t \geq t_0$, the system’s trajectory depends on the initial state $S(t_0)$, input vector $A(t)$, and the probabilistic disturbance terms $W(t)$.
 
Attacker and defender actions exhibit dual input-output roles within the system. At time $t$, the input $A(t) = [A^A(t), A^D(t)]$ comprises the attacker’s action vector $A^a(t) = [a^a_1, a^a_2, \dots, a^a_p]$ and the defender’s action vector $A^D(t) = [a^d_1, a^d_2, \dots, a^d_q]$. The system output $Y(t) = [Y^A(t), Y^D(t)]$ mirrors this structure, with $Y^a(t) = [y^a_1, y^a_2, \dots, y^a_p]$ and $Y^D(t) = [y^d_1, y^d_2, \dots, y^d_q]$. While $p\neq q$ (distinct input-output dimensions for attackers and defenders), these can be standardized via zero-padding to a unified dimension $p'$, where $p' = \max(p, q)$, without loss of generality.
We therefore understand the AutoPT simulation system as a discrete-time system, and its state equation can be represented as a stochastic difference equation:


\begin{equation}
\scriptsize
    \begin{aligned}
        S(t_{k+1})
        =&f\left [S(t_k),A(t_k),W(t_k),t_k\right ]\\
        =&f\left [\left [S^E(t_k),S^A(t_k),S^D(t_k)\right]^T,\left [A^A(t_k),A^D(t_k)\right]^T,W(t_k),t_k\right] \nonumber
    \end{aligned}
\end{equation}

The system's output equation is given as follows:

\begin{equation}
\scriptsize
    \begin{aligned}
        Y(t_{k})
        =&g\left [X(t_k),A(t_k),W(t_k),t_k\right ]\\
        =&g\left [\left [S^E(t_k),S^A(t_k),S^D(t_k)\right]^T,\left [A^A(t_k),A^D(t_k)\right]^T,W(t_k),t_k\right] \nonumber
    \end{aligned}
\end{equation}
where, $t_k$ represents the $k$-th sampling instant within a discrete-time framework, and $k$ serves as the corresponding state index.

Traditional penetration testing research often frames the process as a discrete system driven by isolated attacker and defender actions, modeled as discrete, isolated events~\cite{applebaum2017analysis, miller2018automated, 2020Finding}. However, system outputs are contingent upon structural parameters, system inputs, and initial conditions. This interdependence necessitates a holistic framework that accounts for inherent feedback dynamics, integrating interdependent variables to accurately capture penetration testing environments.



\subsection{Core Modeling Elements}
In this section, we will introduce every core element in detail.

\subsubsection{Network Architecture and Target Assets}

The target network environment involves both the network architecture and target assets. Primarily, network architecture refers to the structural design and configuration of an organization's network systems, which are fundamental in shaping the design of attack vectors and defense strategies during penetration testing. This architecture can be delineated into two levels:

\begin{itemize}
    \item \textbf{Physical Connections:} Physical media, comprising guided media such as cables and optical fibers, and unguided media like wireless signals, facilitate data transmission between network devices. These connections are supported by infrastructure components such as routers, switches, fiber optic systems, network cables, and Bluetooth technology. 

    \item \textbf{Logical Topology:} Logical topology describes the structural relationships and data flow between network devices, such as routers and switches, irrespective of their physical locations or connection methods. It focuses on the interactions and communication pathways within the network, illustrating the interconnection patterns of network assets. Configurations may include star, ring, tree, or hybrid topologies~\cite{Brede2012NetworksAnIM}.
\end{itemize}






Conversely, target assets represent all resources within penetration testing, including hardware, software, data, and personnel. These assets represent the primary objectives that attackers seek to access, manipulate, or compromise, including:

\begin{itemize}
    \item \textbf{Physical Resources:} Servers, workstations, desktops, laptops, mobile devices, external storage, network infrastructure, and security appliances.
    \item \textbf{Virtual Resources:} Operating systems, software applications, open services, databases, and account credentials.
\end{itemize}

The configuration and security posture of these assets determine their value, access policies, and defensive capabilities~\cite{Guo2018Cyberspace}.
Moreover, the modeling of network architecture and target assets can be classified as either static or dynamic. Static modeling assumes that network architecture and configurations remain constant throughout the simulation, whereas dynamic modeling accommodates changes in the target environment due to actions by attackers or defenders, thereby replicating real-world conditions such as:

\begin{itemize}
    \item Variations in the status of enterprise hardware due to employee commuting.
    \item Proactive network changes resulting from dynamic defense strategies like Cyber Mimic Defense~\cite{2016Research} or Moving Target Defense~\cite{2011Moving}.
    \item Randomness and uncertainty in network scenarios, such as the unpredictable failure of network devices and the non-deterministic outcomes of attack actions.
\end{itemize}

\subsubsection{Attacker Models}



The attacker model in AutoPT closely resembles real-world penetration testing methodologies, explicitly delineating identities, targets, and other relevant factors to inform strategic planning, resource allocation, and the selection of attack methods.





\begin{itemize}
    \item \textbf{Identity:} The identity of an attacker may vary from an individual, such as a script kiddie, to a collective, such as a state-sponsored hacking group. This identity could render varying attack methods and available resources. 
    However, in practical testing scenarios, specifying this characteristic is often optional.

    \item \textbf{Objectives:} An attacker’s objectives, influenced by their identity and the specific testing requirements, shape their actions within the network.  Rapid access to sensitive areas prompts the use of direct paths and potent techniques~\cite{hu2020automated}.
   If the objective is to achieve rapid access to sensitive areas, the attacker must employ the shortest attack path and most effective techniques to minimize the steps required. In contrast, if the objective is to identify numerous hidden security vulnerabilities within the network, the attacker should utilize diverse attack methods and carry out a broad spectrum of attacks. If the objective is to maintain a prolonged covert presence, the attacker must implement high-concealment strategies, such as Advanced Persistent Threat, advancing stealthily within the network while meticulously eliminating traces of their movements.
    
    \item \textbf{Actions:} Attacker could exploit vulnerabilities, escalate privileges, conduct scans, and facilitate lateral movement. The Cyber Kill Chain model, developed by Lockheed Martin, offers a systematic approach to penetration testing~\cite{hutchins2011intelligence}. Meanwhile, the MITRE ATT\&CK framework~\cite{2018MITRE} catalogs relevant TTPs (Techniques, Tactics, and Procedures) for AutoPT research. 
    Due to the impracticality of incorporating all TTPs, researchers typically abstract an attack action library based on network architecture, target assets, and attacker-defender models. Each action is defined with specific preconditions, execution steps, and outcomes to optimize the penetration process.
    
    \item \textbf{Network Visibility:} The type of penetration testing—black box, grey box, and white box—influences network visibility. 
    During white box testing, the tester has full visibility and comprehensive knowledge of the system, including access to source code, architectural designs, and network topologies, enabling an in-depth evaluation of internal security measures~\cite{Midian, al2018study, filiol2021method, shravan2014penetration}. 
    Grey box testing involves partial knowledge and access to internal data structures, log files, and application APIs, enhancing testing effectiveness~\cite{goel2015vulnerability, filiol2021method, demott2007revolutionizing}. Conversely, black box testing restricts testers to external observations, simulating real-world attack scenarios~\cite{Midian, awang2013detecting, goel2015vulnerability, al2018study, filiol2021method}. 

\end{itemize}



In AutoPT, the modeling of attackers, particularly their objectives and actions, is essential. The specification of objectives influences agents' tactical decisions and the formulation of evaluation metrics, such as reward functions. Meanwhile, the modeling of actions must capture the variations in attackers' capabilities and realistic impacts while preserving an appropriate level of abstraction. Failure to achieve this balance can complicate the training process and hinder the development of effective strategies.







\subsubsection{Defender Models}
Defenders are the security personnel responsible for safeguarding the network. 
They manage network connections through LANs, VLANs, firewalls, and various security measures~\cite{terranova2024leveraging, furfaro2017using}. Their responsibilities include deploying intrusion detection systems, analyzing logs for anomalies~\cite{ghanem2022towards}, implementing antivirus software, and performing system updates to mitigate vulnerabilities. Additionally, they conduct security awareness programs through lectures and training to achieve protective objectives~\cite{dillon2022perihack}.

Despite their comprehensive visibility of the network, defenders typically lack awareness of the attacker's presence and actions. While they actively monitor network conditions, they may not immediately correlate changes with potential attacks. Based on the level of proactive measures employed, defenders can be categorized into two distinct types:


\begin{itemize}
    \item \textbf{Static Defense:} Static defense is a predefined cybersecurity strategy that operates without adapting to evolving attack policy. This approach is characterized by its fixed nature, activating automatically under specific conditions in response to detected attacker activity or changes in network status. 
    Examples include implementing firewall rules via established logical connections between devices~\cite{schwartz2019autonomous} and ensuring intrusion detection mechanisms via periodic reconnaissance. 
    Within static defense, the success rates of attacker actions can represent defender ability.
    Despite its limitations in immediate response capabilities to new threats and attack techniques, static defense is considered an important auxiliary security measure. Its simplicity and reliability make it a popular choice for AutoPT, as much of the research in this area does not account for defender dynamics.

    \item \textbf{Dynamic Defense:} Dynamic defense involves the definition of  defense agents and an action library, encompassing preconditions, actions, and expected outcomes to enable real-time adjustments to defensive measures based on network conditions. This proactive strategy enhances security by disrupting communications and modifying network information to simulate an active defense~\cite{paul2019learning,elderman2017adversarial}. Furthermore, it complements active defense through methodologies such as Zero Trust Networks, Moving Target Defense, and honeypots. The flexibility inherent in dynamic defense modeling is essential for addressing the evolving landscape of cyber threats, ensuring effective protection by considering both attacker behavior and changes in network environments.

\end{itemize}

Upon establishing the core elements, penetration testing can be understood as a dynamic interaction between attackers and defenders or as a series of unidirectional offensive maneuvers by attackers. 
The target network's state evolves based on predefined logical relationships that dictate the outcomes of various actions. Furthermore, routine user activities may also alter the network's assets and topology~\cite{applebaum2017analysis}. These changes collectively influence the decision-making for subsequent attack and defense actions.  The penetration test is conducted through the iterative application of these strategies throughout the simulation.
Our modeling classification method is constructed based on three perspectives: the overall aim of the research literature, the key elements presented in this section, and the interactions between these elements.
We will first present the overall contributions of the paper, followed by an analysis on the core modeling factors.

\section{The Multi-Dimensional Classification System for Penetration Testing Modeling: \modelcla}

\begin{figure*}[tb]
    \centering
    \includegraphics[width=130mm]{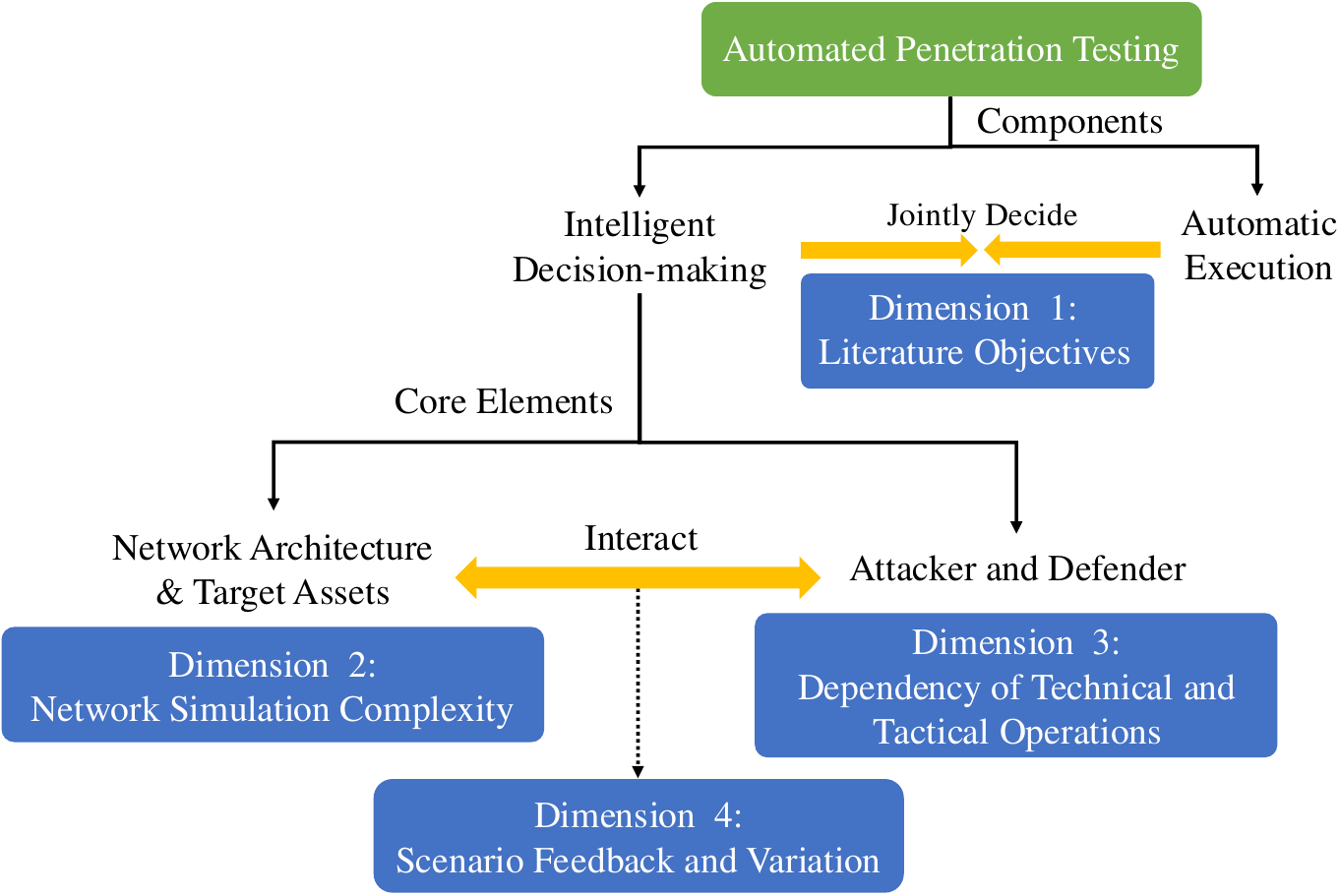}
    \caption{The Multi-Dimensional
Classification System for Penetration Testing Modeling}
    \label{Class_dim}
\end{figure*}

In this section, we present the \textbf{M}ulti-\textbf{D}imensional \textbf{C}lassification System for \textbf{P}enetration Testing \textbf{M}odeling (\modelcla), an innovative approach for classifying target network scenarios modeling method in penetration testing. 
This system categorizes scenario modeling based on four principal dimensions: (1) Literature Objectives, (2) Network Simulation Complexity, (3) Dependency of Technical and Tactical Operations, and (4) Scenario Feedback and Variation. Each dimension comprises sub-dimensions for a nuanced classification of attack scenarios, as illustrated in Figure~\ref{Class_dim}. This paper begins with an explanation of our classification criteria, followed by illustrative examples, and concludes with an analysis of the reviewed literature.

\subsection{Dimensionality of~\modelcla}
In this section, we define and introduce the four primary dimensions characterizing the \modelcla, along with their associated sub-dimensions.






\subsubsection{Literature Objectives}


The Literature Objectives analyze the aims and outcomes of various AutoPT studies, including tool development, policy proposals, and platform introductions. Based on the two phases of AutoPT—intelligent decision-making and automatic execution—we categorize this dimension into three sub-dimensions: . 
Specifically, technical automation pertains to automatic execution, while policy automation refers to intelligent decision-making; when a study simultaneously addresses both aspects, we classify it as complete automation.
By organizing the research background, objectives, and significance of each study, we identify the literature's goals and gain insights into each paper's contributions, innovations, and applications.


\textbf{Technical Automation:} This dimension emphasizes the direct implementation of specific technical and tactical procedures without strategic planning. It represents the earliest stage of automation, characterized by a lower level of intelligence in AutoPt.
It automates predefined testing technicals such as identifying live hosts, open ports, services, and conducting vulnerability scans. Tools are central to this process. Nmap~\cite{lyon2009nmap}, Fscan~\cite{Fscan}, and Webshell~\cite{longxiao2018webshell} facilitate the identification of live hosts and services. Nessus and AWVS conduct automated vulnerability scans. Metasploit automates vulnerability verification and exploitation. These tools are essential for the autonomous execution phase of AutoPT, enabling the efficient performance of repetitive tasks and minimizing manual errors and time expenditure.

\textbf{Policy Automation:} This aspect focuses on automated generation of attack policy without real-world execution, crucial for intelligent decision-making in AutoPT. It involves the automated planning of attack paths and technical actions, representing a key focus in contemporary AI research for AutoPT~\cite{enoch2020harmer,schwartz2020pomdp+,becker2024evaluation}. For instance, Hu et al. use Multi-host Multi-stage Vulnerability Analysis to construct an attack tree for network topologies, applying Deep Q-Networks (DQN) to identify the most exploitable attack paths~\cite{hu2020automated}. Zhou et al.~\cite{zhou2021autonomous} frame penetration testing as a Markov Decision Process and used an improved deep q-network to decouple actions and learn attack strategies. However, their methodologies remain limited to policy generation in theoretical network environments, lacking integration with actual penetration tools or execution of attack payloads in real-world scenarios.


\textbf{Complete Automation:} 
While the two aforementioned aspects are crucial, they both represent only singular aspects of penetration testing. A comprehensive security assessment requires the integration of automatic execution and intelligent decision-making to effectively identify and mitigate potential security threats.
Complete automation encompasses automation of the entire attack lifecycle, from decision-making to execution. This includes the automatic planning of attack paths and the integration of execution tools with actual payloads to perform real-world penetration tests in either simulated or live network environments, entirely without human intervention~\cite{dorchuck2021goal,sarraute2013automated,xu2024autoattacker}.

Literature objectives deepen our understanding of research trends in AutoPT, providing insights into theoretical advancements and practical implications while clarifying the current landscape and future directions. 
Moreover, these objectives could also facilitate the classification of network simulation complexity, modeling dependencies between technical and tactical operations, as well as scenario feedback and variation.




\subsubsection{Network Simulation Complexity}

This dimension focuses on network architecture and target assets-the first element of AutoPT simulation modeling. This dimension is further divided into two sub-dimensions based on the abstraction level and construction methods of network attributes: hypothetical and authentic attributes. 

\textbf{Simulation of Hypothetical Attributes:} Numerous studies utilize numerical, rule-based, or conceptual methods to abstractly model assets and architectures. For instance, Hammar et al. utilize numerical attributes to characterize nodes in a four-node network, with each node represented by multidimensional metrics of defensive and detection capabilities~\cite{2020Finding}. 




\textbf{Simulation of Authentic Attributes:} Certain studies employ real-world systems, software, services, account passwords, vulnerabilities, and other real information to model target assets and attributes. These works utilize complex network topologies to accurately replicate real-life environments, reflecting both node attributes and their interrelationships. For example, Microsoft's CyberBattleSim defines the operating systems, software, vulnerabilities, and node reward for each node while establishing diverse connection relationships across multiple small scenarios involving fewer than 20 nodes~\cite{Cyberbattlesim}.





\subsubsection{Dependency of Technical and Tactical Operations}

This dimension analyzes attacker and defender models to assess whether their defined actions incorporate interdependencies, specifically determining if the outcome of one action serves as a prerequisite for subsequent actions.

\textbf{Isolated Technical and Tactical Actions:} Many technical and tactical actions are independent, lacking defined prerequisites. This category includes executing single tactics, such as using Nmap for scanning, or multiple unrestricted actions.
For example, Sarraute et al.~\cite{sarraute2013penetration} defined only scanning and vulnerability exploitation actions in a penetration scenario. Although agents may implicitly learn that scanning before exploitation improves the success rate of vulnerability selection, the action definitions do not specify prerequisites, allowing actions to be performed independently.

\textbf{Coordinated Technical and Tactical Actions:} Many technical and tactical actions are interdependent, forming an integrated kill chain. 
This coordination is evident in stages such as privilege escalation and lateral movement, where initial actions like vulnerability exploitation or phishing attacks precede subsequent activities such as credential theft and malware implantation. Defining the preconditions and post-effects of actions is essential. For example, Filiol et al.~\cite{filiol2021method} modeled attacker behavior by specifying logical relationships among actions, including domain name acquisition, IP scanning, service version collection, attack list generation, and attack configuration.


Dependency of Technical and Tactical Operations investigates the explicit interrelationships between sequential actions essential for precise simulation modeling. 
In real-world penetration testing scenarios, actions are constrained by execution limitations, and their variability hinders standardization efforts.
Many studies overlook these dependencies, thereby simplifying execution constraints and standardizing decision parameters, which increases abstraction and reduces alignment with real-world conditions. Addressing this gap is crucial for the effective transition of simulation models to practical environments.

In existing research, isolated and coordinated technical and tactical operations are frequently combined. For example, CALDERA ~\cite{applebaum2016intelligent} employs coordinated technical and tactical actions to simulate an attacker’s lateral movement, privilege escalation, and data theft by imposing action dependencies. Meanwhile, it also utilizes isolated technical and tactical actions to represent a passive defender unaware of the attacker, limited to independent actions such as random reboots and logins. Although classified as a continuous tactic scenario, CALDERA incorporates isolated tactics in its modeling.

 



\subsubsection{Scenario Feedback and Variation}


This dimension classifies modifications to the target network's architecture and assets, including changes in host connectivity, installed systems and software, account credentials, and vulnerabilities. It does not account for attributes related to attackers and defenders, such as an attacker’s privilege level or newly acquired credentials.
This dimension involves two types of changes: scenario feedback and scenario variation. 

\begin{figure}[tb]
    \centering
    \includegraphics[width=80mm]{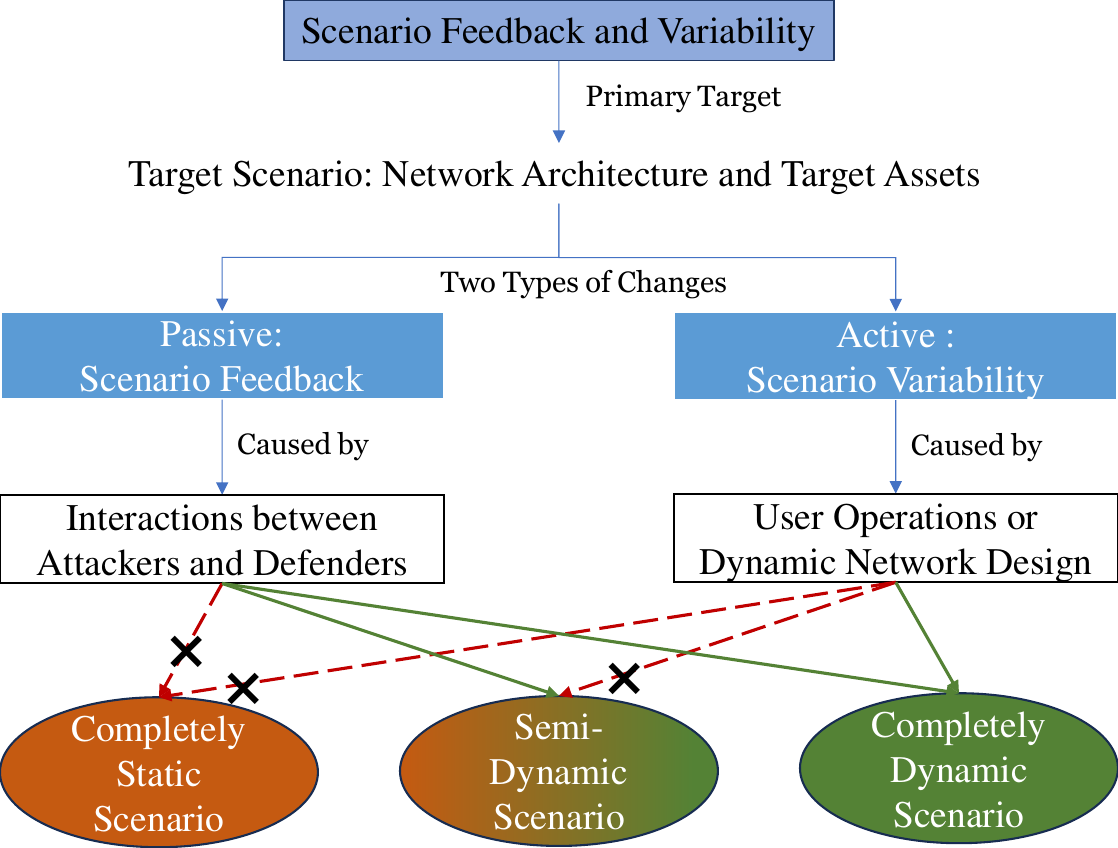}
    \caption{The connotation and sub-dimensions of the Scenario Feedback and Variation}
    \label{the_forth_dim}
\end{figure}

\textbf{Scenario Feedback} refers to passive changes arising from interactions between attackers and defenders that affect the target network's architecture and assets. This includes attacker actions such as establishing connections, deploying phishing emails or malware, and causing network disruptions, as well as defender responses like system shutdowns, credential remediation, and software updates. These alterations occur only when attackers and defenders engage, characterizing them as passive changes.
In contrast, \textbf{Scenario Variation} involves predefined modifications within simulation models designed to emulate real-world user operations or dynamic network configurations. Examples include simulating user behavior, scheduling power operations to reflect work routines, conducting traffic simulations for behavioral drills, periodically updating IP addresses and systems, and implementing defense strategies such as Moving Target Defense (MTD), Cyber Mimic Defense (CMD), load balancing, and Endogenous Safety and Security (ESS). These changes are integrated into the scenario and execute according to predetermined schedules or conditions, independent of the immediate actions of attackers or defenders, thus qualifying them as active changes.
Figure~\ref{the_forth_dim} illustrates the Scenario Feedback and Variation dimension.

Based on these two aspects in AutoPT modeling, scenario feedback and variation are categorized into three sub-dimensions:

\textbf{Completely Static Scenario:} A scenario with no passive modifications from attack-defense interactions and no active alterations to the target environment.

\textbf{Semi-Dynamic Scenario:} A scenario that incorporates passive changes resulting from attack-defense interactions but does not include active modifications to the target environment.

\textbf{Completely Dynamic Scenario:} A scenario that encompasses both passive changes from attack-defense interactions and active alterations to the target environment.


\subsection{Case Study}

To clarify~\modelcla, we present and analyze four representative cases, including:

\begin{itemize}
    \item Penetration Testing Tools: These tools automate key security tactics and techniques, essential for conducting AutoPT in real-world scenarios.
    \item Numerical Simulation Networks~\cite{2020Finding}: Although highly abstract and detached from real-world conditions, these networks provide a theoretical framework for exploring cybersecurity complexities.
    \item CyberBattleSim~\cite{Cyberbattlesim}: Developed by Microsoft, this simulation platform is employed in various studies for both simulation and emulation purposes.
    \item Network Attack Simulator~\cite{schwartz2019autonomous}: Released by Harvard University, this popular simulator specializes in network attack simulations, enhancing the understanding of penetration testing behaviors.
\end{itemize}


These cases are widely cited and diverse, encompassing a broad spectrum of classifications in existing research. Analyzing them provides deeper insights into the classification principles and applications of \modelcla.

\subsubsection{Penetration Testing Tools}

Penetration testing tools typically incorporate technical automation, simulate authentic attributes, execute isolated technical and tactical actions, and utilize either entirely static scenarios or impose no scenario restrictions. These AutoPT tools are essential for network security, enabling professionals to identify and evaluate vulnerabilities within networks, applications, and systems.
The following are some widely used automated tools:

\begin{itemize}
    \item Nmap (Network Mapper)~\cite{lyon2009nmap}: A multifunctional security and port scanner designed to efficiently evaluate individual hosts or large networks. It offers features such as host discovery, port scanning, service identification, operating system detection, version scanning, and script scanning.
    \item Nessus~\cite{TenableNessus}: A comprehensive vulnerability scanner developed by Tenable, designed to identify security vulnerabilities in systems, networks, and applications. It scans targets such as IP addresses and domains, and generates detailed reports that include vulnerability descriptions, severity ratings, and recommended remediation actions.
    \item Metasploit~\cite{MetasploitWebsite}: Metasploit is a collaborative framework designed for vulnerability validation and security assessments, operating through distinct modules. Auxiliary Modules perform scanning, fingerprinting, e.t.c, to support penetration testing. Exploit Modules utilize identified vulnerabilities to infiltrate systems. Payload Modules execute post-exploitation tasks, enabling arbitrary command execution on targets. Post-Exploitation Modules secure further access and gather additional data from compromised systems. Encoder Modules obfuscate payloads to bypass security mechanisms.

\end{itemize}


Many AutoPT tools automate specific penetration testing steps but often require manual input for targets and parameters, making it difficult to conduct the entire process without human involvement. These tools fall under Technical Automation and simulate authentic attributes based on real-world network. Their technical and tactical measures can be executed in isolation, categorizing them as Isolated Technical and Tactical Actions. Although applicable to various scenarios, they are typically confined to secure environments due to legal and ethical considerations and operate within static scenarios without active user interaction.






\subsubsection{Numerical Simulation Networks}
Numerical simulation networks encompass policy automation, simulation of hypothetical attributes, execution of isolated technical and tactical actions, and semi-dynamic scenarios. 

\begin{figure}[tb]
    \centering
    \includegraphics[width=85mm]{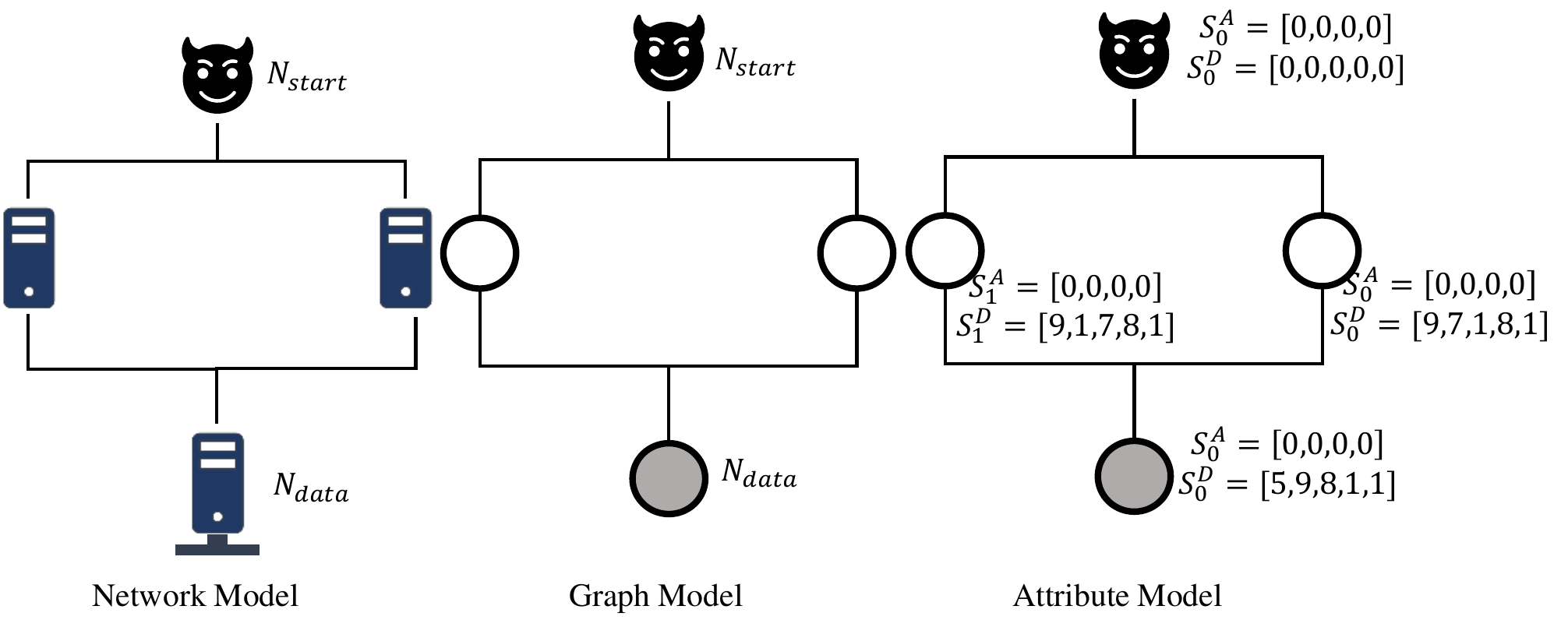}
    \caption{A numerical simulation network with four nodes~\cite{2020Finding}. The left, middle, right diagram shows the network model, graph model and attribute model, respectively.}
    \label{fig3:four nodes}
\end{figure}

Hammer et al. investigated attack–defense interactions in penetration testing using a four-node numerical simulation network~\cite{2020Finding}. Figure~\ref{fig3:four nodes} presents the network architecture, its graphical representation and attribute model.
In the left diagram, $N_{start}$ represents the attacker's computer, while the other nodes correspond to defender components. The attacker’s objective is to compromise $N_{data}$. The middle diagram formalizes the network as a graph, with nodes representing components and edges indicating connections. 
Each node $k$ is characterized by attributes $S_k = \{S^A_k, S^D_k\}$, which include both attack and defense values. The attack attributes $S^A_k = \{S^A_{k,1}, S^A_{k,2}, \ldots, S^A_{k,m}\}$ represent the strength of $m$ attack types and are visible only to the attacker. The defense attributes $S^D_k = \{S^D_{k,1}, S^D_{k,2}, \ldots, S^D_{k,m+1}\}$ are visible only to the defender, where the first $m$ attributes correspond to the respective attack types and the $(m+1)$-th attribute indicates detection capability. This study simulates hypothetical attributes, and the target network remains static.

The attacker can perform two actions on a node $k$: (1) reconnaissance to reveal the defense state $S^D_k$, and (2) execute an attack of type $j \in \{1,2,\cdots, m\}$, increasing the attack state $S^A_{k,j}$ by one. The defender can take two actions on node $k$: (1) monitoring operations to enhance the node's detection ability $S^D_{k;m+1}$, and (2) defensive operations to strengthen defenses against attack type $j \in \{1,2,\cdots, m\}$, thereby increasing $S^D_{k,j}$. The attacker and defender alternate actions. If $S^A_{k,j} > S^D_{k,j}$ for any attack type $j$, the attacker compromises node $k$, making its neighbors visible. If the attack does not compromise the node, the defender detects it with probability $p = \frac{S^D_{k;m+1}}{w+1}$, based on the node's detection ability $S^D_{k;m+1}$.

From the attacker and defender models, it is evident that both can execute actions without constraints, allowing them to act independently. This categorizes their actions as Isolated Technical and Tactical Actions. The defender’s actions modify the network’s defense attributes and enhance its monitoring capabilities, resulting in passive changes to the network. Since there are no active alterations, the scenario is classified as Semi-Dynamic. The game aims to determine optimal strategies without involving real-world tactical executions or attack payloads, classifying it as Policy Automation.

In summary, Hammer et al.’s approach is characterized by Policy Automation, Simulation of Hypothetical Attributes, Isolated Technical and Tactical Actions, and a Semi-Dynamic Scenario.

\subsubsection{CyberBattleSim}
CyberBattleSim is an open-source research project initiated by Microsoft in 2021 that uses high-level abstractions of computer networks and cybersecurity concepts to study how autonomous agents operate within simulated corporate environments~\cite{Cyberbattlesim}. Numerous studies have used it for AutoPT research~\cite{li2024knowledge, terranova2024leveraging, zhang2022improved, guo2023automated}. With our classification system, CyberBattleSim is categorized under policy automation, simulation of authentic attributes, coordinated technical and tactical actions, and semi-dynamic scenarios. 

\begin{figure}[tb]
    \centering
    \includegraphics[width=85mm]{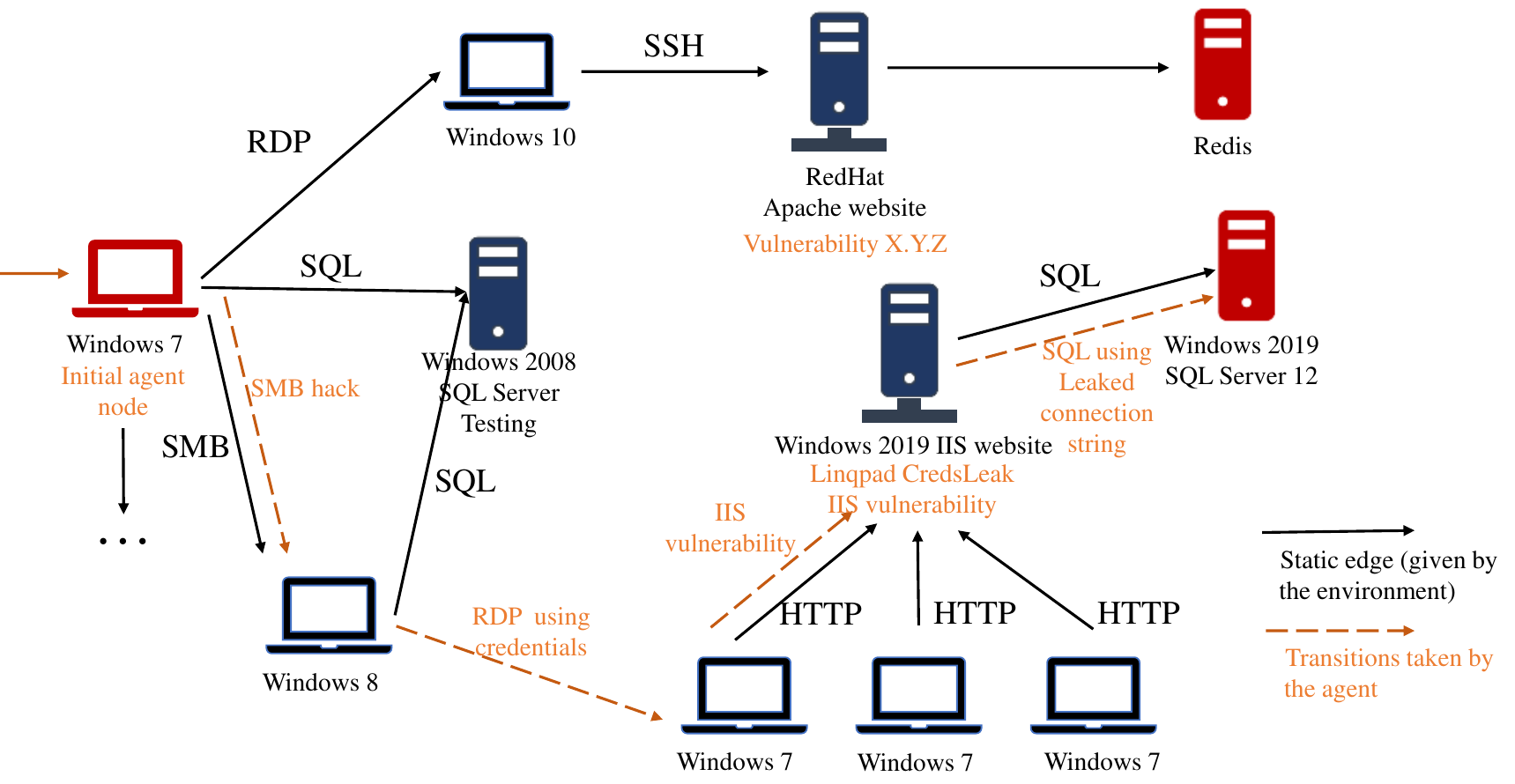}
    \caption{Schematic Diagram of CyberBattleSim Network~\cite{Cyberbattlesim}}
    \label{CyberBattleSim}
\end{figure}

CyberBattleSim focuses on threat modeling during the post-compromise lateral movement phase of network attacks. It simulates a fixed network topology with parameterized vulnerabilities, allowing attackers to exploit these weaknesses for lateral movement. A target network scenario, illustrated in Figure~\ref{CyberBattleSim}, consists of nodes running various operating systems and software. Each computer has specific attributes, values, and pre-assigned vulnerabilities. Communication between nodes is depicted by black edges labeled with communication protocols. The target network is constructed using realistic attribute simulations, and the scenario remains static without active changes.


The attacker aims to gain network control by exploiting vulnerabilities and maximizing rewards through three actions: performing a local attack, performing a remote attack, and connecting to other nodes. Actions are parameterized by the source node where the underlying operation should take place, and they are only permitted on nodes owned by the agent. As illustrated in Figure~\ref{CyberBattleSim}, the attacker starts from a Windows 7 node, exploits vulnerabilities, and uses cached credentials to move laterally, ultimately accessing an SQL database. Defenders monitor activities to detect and mitigate attacks by reimaging infected nodes. Attack success also depends on predefined probabilities.

The attacker modifies network communication and architecture, while defenders reimage systems, patch vulnerabilities, and alter node attributes. Consequently, the target network undergoes passive changes from both attacker and defender actions without active alterations, classifying it as a semi-dynamic scenario. CyberBattleSim conducts abstract simulations without executing real attack code, emphasizing agent interactions. It is designed for small to medium networks (10–20 nodes) and does not support fully dynamic scenarios.





\subsubsection{Network Attack Simulator}

\begin{figure}[tb]
\centering
\includegraphics[width=85mm]{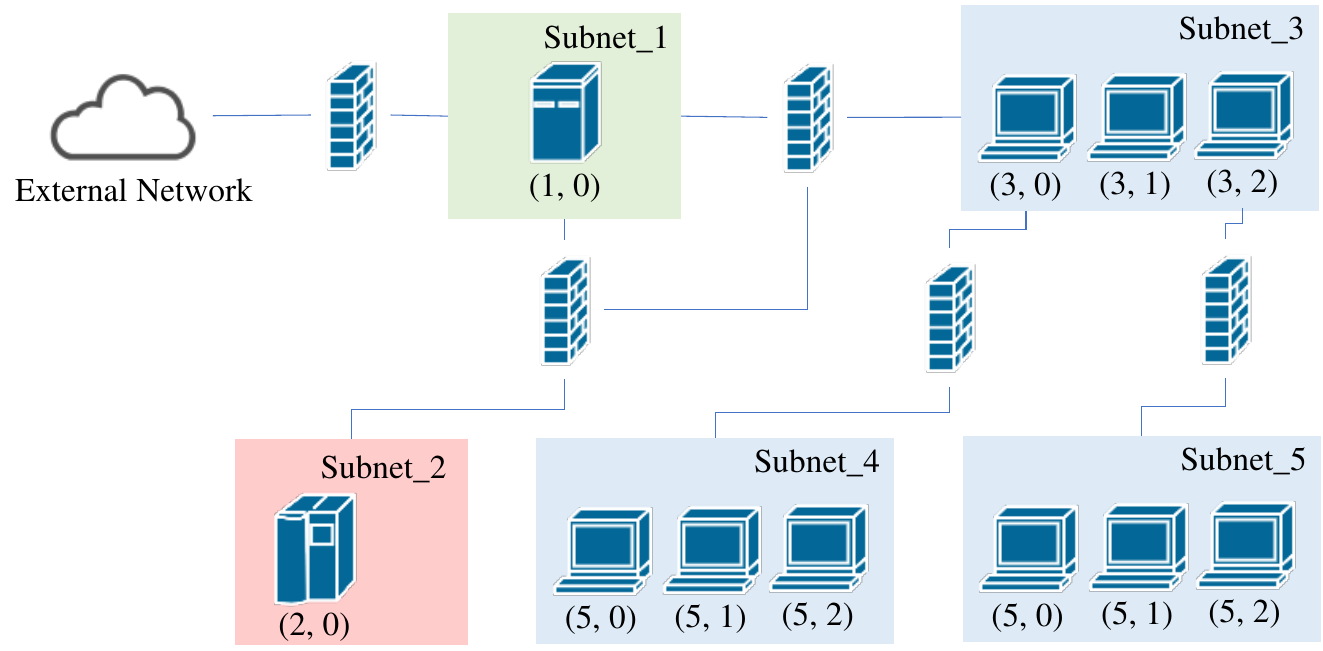}
\caption{Network Architecture Diagram of Network Attack Simulator~\cite{schwartz2019autonomous}}
\label{Network Attack Simulator}
\end{figure}

The Network Attack Simulator~\cite{schwartz2019autonomous}, a lightweight, open-source tool developed by Schwartz et al. in 2019, is a groundbreaking application of reinforcement learning in AutoPT research.
The simulator constructs a network environment of multiple subnets with firewall-controlled access, each containing machines running various services. As shown in Figure~\ref{Network Attack Simulator}~\cite{schwartz2019autonomous}, the architecture includes node attributes such as address (subnet\_ID, machine\_ID), machine value, and parameters (open services, success rate, exploitation cost).
The network architecture and asset modeling utilize real-world data for the simulation of authentic attributes, remaining static throughout the process.

\begin{table*}[b!]
\centering
\renewcommand\arraystretch{1}
\caption{Classification Table of Simulation Modeling Methods in Typical Literature on Automated Penetration Testing}
\label{Classification_Table}
\resizebox{\linewidth}{!}{
\begin{tabular}{c|c|ccc|cc|cc|ccc}
\toprule[2pt]
\multirow{2}{*}{Year} & \multirow{2}{*}{Paper} & \multicolumn{3}{c|}{Literature Objectives}         & \multicolumn{2}{c|}{Network Simulation Complexity}         & \multicolumn{2}{c|}{Dependency of T\&T Operations}         & \multicolumn{3}{c}{Scenario Feedback and Variation}                \\ \cline{3-12}
& & Technical & Policy & Complete & Hypothetical & Authentic & Isolated & Coordinated & Static & Semi-Dynamic & Dynamic \\ 
\midrule[1.5pt]
\multirow{2}{*}{1997} & Haeni et al.~\citep{haeni1997firewall}         & \checkmark &   &   &   & \checkmark & \checkmark &   & — & — & — \\
&Nmap \citep{nmap}  & \checkmark &   &   &   & \checkmark & \checkmark &   & — & — & — \\
\hline

1998 &Nessus~\citep{TenableNessus}             & \checkmark &   &   &   & \checkmark & \checkmark &   & — & — & — \\
 \hline
2001 &McDermott et al.~\citep{mcdermott2001attack}       &   & \checkmark &   & \checkmark &   &   & \checkmark & \checkmark                &                  &                  \\
 \hline
2002 &Skaggs et al.~\citep{skaggs2002network}         & \checkmark &   &   &   & \checkmark & \checkmark &   & \checkmark                &                  &                  \\ \hline
2003 &Metasploit~\citep{MetasploitWebsite}         & \checkmark &   &   &   & \checkmark & \checkmark & & — & — & — \\ \hline
2005 &Liu et al.~\citep{liu2005game}               &   & \checkmark &   &   & \checkmark & \checkmark &   & \checkmark                &                  &                  \\ \hline
\multirow{4}{*}{2007}  &Kosuga et al.~\citep{kosuga2007sania}           & \checkmark &   &   &   & \checkmark & \checkmark &   & \checkmark                &                  &                  \\
 &Fonseca et al.~\citep{fonseca2007testing}        & \checkmark &   &   &   & \checkmark & \checkmark &   & \checkmark                &                  &                  \\
 &Shen et al.~\citep{shen2007strategies}        &   & \checkmark &   &   & \checkmark &   & \checkmark &                  & \checkmark                &                  \\
 &Cone et al.~\citep{cone2007video}             &   & \checkmark &   & \checkmark &   & \checkmark &   & \checkmark                &                  &                  \\ \hline
\multirow{2}{*}{2009}   &Lyon et al.~\citep{lyon2009nmap}              & \checkmark &   &   &   & \checkmark & \checkmark &   & — & — & — \\
 &Greenwald et al.~\citep{greenwald2009automated}    &   & \checkmark &   &   & \checkmark & \checkmark &   & \checkmark                &                  &                  \\ \hline
2011 &Sarraute et al.~\citep{sarraute2011algorithm}     &   & \checkmark &   &   & \checkmark &   & \checkmark &                  & \checkmark                &                  \\ \hline
\multirow{3}{*}{2013} 
 &Sarraute et al.~\citep{sarraute2013penetration}   &   & \checkmark &   &   & \checkmark & \checkmark &   & \checkmark                &                  &                  \\
 &Sarraute et al.~\citep{sarraute2013automated}     &   &   & \checkmark &   & \checkmark &   & \checkmark &                  & \checkmark                &                  \\
 &Van Dijk et al.~\citep{van2013flipit}             &   & \checkmark &   & \checkmark &   & \checkmark &   & \checkmark                &                  &                  \\  \hline
2014 &Chapman et al.~\citep{chapman2014playing}        &   & \checkmark &   & \checkmark &   & \checkmark &   & \checkmark                &                  &                  \\ \hline
\multirow{2}{*}{2016}  &Applebaum et al.~\citep{applebaum2016intelligent}  &   & \checkmark &   & \checkmark &   & \checkmark &   &                  & \checkmark                &                  \\ 
 &Chapman et al.~\citep{chapman2016cyber}          &   & \checkmark &   & \checkmark &   & \checkmark &   & \checkmark                &                  &                  \\  \hline
\multirow{3}{*}{2017}  &Elderman et al.~\citep{elderman2017adversarial}   &   & \checkmark &   & \checkmark &   & \checkmark &   &                  & \checkmark                &                  \\
 &Applebaum et al.~\citep{applebaum2017analysis}     &   & \checkmark &   &   & \checkmark &   & \checkmark &                  &                  & \checkmark                \\
 &Ficco et al.~\citep{ficco2017simulation}       & \checkmark &   &   &   & \checkmark & \checkmark &   & \checkmark                &                  &                  \\ \hline
\multirow{4}{*}{2018}   &Miller et al.~\citep{miller2018automated}       &   & \checkmark &   &   & \checkmark &   & \checkmark & \checkmark                &                  &                  \\
 &Ghanem et al.~\citep{ghanem2018reinforcement}   &   & \checkmark &   &   & \checkmark &   & \checkmark & \checkmark                &                  &                  \\
 &Casola et al.~\citep{casola2018towards}         & \checkmark &   &   &   & \checkmark &   & \checkmark & \checkmark                &                  &                  \\
 &Ghanem et al.~\citep{ghanem2019reinforcement}   &   &  &  \checkmark &   & \checkmark &   & \checkmark &                  & \checkmark                &                  \\ \hline
\multirow{6}{*}{2019} &Paul et al.~\citep{paul2019learning}          &   & \checkmark &   &   & \checkmark &   & \checkmark &                  & \checkmark                &                  \\
 &Schwartz et al.~\citep{schwartz2019autonomous}    &   & \checkmark &   &   & \checkmark &   & \checkmark & \checkmark                &                  &                  \\
 &Paul et al.~\citep{paul2019strategic}         &   & \checkmark &   & \checkmark &   & \checkmark &   &                  & \checkmark                &                  \\
 &Zhou et al.~\citep{2019NIG}                   &   & \checkmark &   &   & \checkmark &   & \checkmark & \checkmark                &                  &                  \\
 &Zang et al.~\citep{yichao2019improved}        &   & \checkmark &   &   & \checkmark &   & \checkmark &                  & \checkmark                &                  \\ \hline
\multirow{12}{*}{2020}  &Hu et al.~\citep{hu2020automated}           &   & \checkmark &   &   & \checkmark &   & \checkmark & \checkmark                &                  &                  \\
 &Hammar et al.~\citep{2020Finding}         &   & \checkmark &   & \checkmark &   & \checkmark &   &                  & \checkmark                &                  \\
 &Valea et al.~\citep{valea2020towards}          & \checkmark &   &   &   & \checkmark & \checkmark &   & \checkmark                &                  &                  \\
 &Bhattacharya et al.~\citep{bhattacharya2020automated} &   & \checkmark &   &   & \checkmark &   & \checkmark &                  & \checkmark                &                  \\
 &Nguyen et al.~\citep{nguyen2020multiple}        &   & \checkmark &   &   & \checkmark & \checkmark &   & \checkmark                &                  &                  \\
 &Costa et al.~\citep{costa2020charles}          & \checkmark &   &   &   & \checkmark & \checkmark &   & \checkmark                &                  &                  \\
 &Chowdhary et al.~\citep{chowdhary2020autonomous}   &   & \checkmark &   &   & \checkmark &   & \checkmark & \checkmark                &                  &                  \\
 &Bland et al.~\citep{bland2020machine}          &   & \checkmark &   &   & \checkmark &   & \checkmark &                  & \checkmark                &                  \\
 &Hu et al.~\citep{hu2020apu}                 &   & \checkmark &   &   & \checkmark & \checkmark &   & \checkmark                &                  &                  \\
 &Enoch et al.~\citep{enoch2020harmer}           &   & \checkmark &   &   & \checkmark &   & \checkmark & \checkmark                &                  &                  \\
 &Schwartz et al.~\citep{schwartz2020pomdp+}        &   & \checkmark &   &   & \checkmark & \checkmark &   &                  & \checkmark                &                  \\ \hline
 \multirow{7}{*}{2021}  
 &Dorchuck et al.~\citep{dorchuck2021goal}          &   &   & \checkmark &   & \checkmark &   & \checkmark & \checkmark                &                  &                  \\
 &Qian et al.~\citep{qian2021ontology}          &   & \checkmark &   &   & \checkmark &   & \checkmark & \checkmark                &                  &                  \\
 &Filiol et al.~\citep{filiol2021method}          & \checkmark &   &   &   & \checkmark &   & \checkmark & \checkmark                &                  &                  \\
 &Zhou et al.~\citep{zhou2021autonomous}        &   & \checkmark &   &   & \checkmark &   & \checkmark & \checkmark                &                  &                  \\
 &Hacks et al.~\citep{hacks2021towards}          &   & \checkmark &   &   & \checkmark &   & \checkmark &                  & \checkmark                &                  \\
 &Erd{\H{o}}di et al.~\citep{erdHodi2021simulating}     & \checkmark &   &   &   & \checkmark &   & \checkmark & \checkmark                &                  &                  \\
 &Ji et al.~\citep{ji2021optimal}             &   & \checkmark &   &   & \checkmark & \checkmark &   &                  & \checkmark                &                  \\ \hline
\multirow{5}{*}{2022}  
 &Dillon et al.~\citep{dillon2022perihack}        &   & \checkmark &   & \checkmark &   &   & \checkmark &                  & \checkmark                &                  \\
 &Yamin et al.~\citep{yamin2022use}              &   &   & \checkmark &   & \checkmark &   & \checkmark &                  & \checkmark                &                  \\
 &Confido et al.~\citep{confido2022reinforcing}    &   & \checkmark &   &   & \checkmark &   & \checkmark &                  & \checkmark                &                  \\
 &Tran et al.~\citep{tran2022cascaded}          &   & \checkmark &   &   & \checkmark &   & \checkmark &                  & \checkmark                &                  \\
 &Hance et al.~\citep{hance2022distributed}      & \checkmark &   &   &   & \checkmark &   & \checkmark & \checkmark                &                  &                  \\ \hline
\multirow{2}{*}{2023}   &F{\ae}r{\o}y et al.~\citep{faeroy2023automatic}       & \checkmark &   &   &   & \checkmark & \checkmark &   & \checkmark                &                  &                  \\
 &Li et al.~\citep{li2023innes}                    &   & \checkmark &   &   & \checkmark &   & \checkmark & \checkmark                &                  &                  \\ \hline
\multirow{8}{*}{2024}   &Xu et al.~\citep{xu2024autoattacker}        &   &   & \checkmark &   & \checkmark &   & \checkmark & \checkmark                &                  &                  \\
 &Becker et al.~\citep{becker2024evaluation}      &   & \checkmark &   &   & \checkmark &   & \checkmark & \checkmark                &                  &                  \\
 &Li et al.~\citep{li2024knowledge}           &   & \checkmark &   &   & \checkmark &   & \checkmark &                  & \checkmark                &                  \\
 &Alshehri et al.~\citep{alshehri2024breachseek}    & \checkmark &   &   &   & \checkmark &   & \checkmark & \checkmark                &                  &                  \\
 &Deng et al.~\citep{deng2024pentestgpt}        &   & \checkmark &   &   & \checkmark &   & \checkmark & \checkmark                &                  &                  \\
 &Wang et al.~\citep{Zhenduo}                   &   & \checkmark &   &   & \checkmark &   & \checkmark & \checkmark                &                  &                 \\
 &Li et al.~\citep{li2024dynpen} &   & \checkmark &   &   & \checkmark &  \checkmark &  &                 &                  &  \checkmark               \\
 &Wang et al.~\citep{wang2024pentraformer} &   & \checkmark &   &   & \checkmark &  \checkmark &  & \checkmark                &                  &                 \\

    \bottomrule[2pt]
    \end{tabular}
}
\end{table*}

Focusing solely on attacker modeling, the simulator allows for individual scanning actions and vulnerability exploitation targeting services on each machine. Scanning identifies services on ports, which are then exploited based on the machine's configuration. The attacker's actions are independent; one does not depend on the completion of another. Although scanning can guide vulnerability selection, traversing execution vulnerabilities can also grant access to the target machine. Importantly, these actions do not alter the target network, reinforcing the classification of the scenario as completely static.

In summary, Network Attack Simulator combines policy automation, authentic attribute simulation, isolated technical and tactical actions, and a completely static scenario. While it implicitly represents static defenders through subnet and machine connections and vulnerability success rates, the tool's simplicity limits its scalability for larger networks and lacks explicit defender modeling or a dynamically changing network environment.





\subsection{Research on Existing Penetration Testing Scenario Modeling Methods}

We conducted a systematic literature review of AutoPT studies using Web of Science, Scopus, and IEEE Xplore (1990s-present). A two-stage screening process filtered out irrelevant and low-quality papers. Inclusion criteria consisted of thematic relevance, methodological rigor, academic impact, and research recency.
After applying these criteria, 65 representative documents were selected for analysis, with 33 from 2020-2024. Each study was cross-reviewed by at least two researchers to ensure accuracy and reliability. We categorized AutoPT modeling methods according to their characteristics (Table~\ref{Classification_Table}) and examined the research background, objectives, and significance of each article.


We summarized the article count across four dimensions in Table~\ref{Classification Statistics Table}. Notably, some automated execution tools like Nessus, Metasploit, and nmap have unrestricted application scenarios (therefore denoted by dash notation). Generally, their use requires consideration of legal and ethical constraints, typically in isolated network environments, categorized as Completely Static Scenarios.

\begin{table*}[h!t]
\centering
\caption{Classification Statistics of Simulation Modeling Methods in Automated Penetration Testing Literature}
\label{Classification Statistics Table}
\resizebox{\linewidth}{!}{
\begin{tabular}{c|ccc|cc|cc|ccc}
\toprule[2pt]
\multirow{2}{*}{Types} &\multicolumn{3}{c|}{Literature Objectives}         & \multicolumn{2}{c|}{Network Simulation Complexity}         & \multicolumn{2}{c|}{Dependency of T\&T Operations}         & \multicolumn{3}{c}{Scenario Feedback and Variation}                \\ \cline{2-11}
& Technical &   Policy & Complete  & Hypothetical  &  Authentic & Isolated & Coordinated & Completely Static& Semi-Dynamic &  Completely Dynamic  \\ 
\hline


Quantity& 17 & 43 & 5 & 10 & 55 & 29 & 36 & 43 & 20 & 2\\
\bottomrule[2pt]
\end{tabular}
}
\end{table*}

Policy automation and intelligent decision-making are prominent research areas, attracting significant academic attention. Most studies focus on simulating authentic attributes and continuous technical actions, closely mirroring practical scenarios. However, research on dynamic environments remains limited. For example, Applebaum et al.~\cite{applebaum2017analysis} introduce active network changes using gray agents, but initiate only one connection set per round, resulting in minimal alterations within small to medium networks (11–21 hosts). Similarly, Li et al.~\cite{li2024dynpen} carefully define network changes but do not quantify them or test scalability in larger networks. Their experiments are confined to a 10-node network and overlook dynamic simulation and emulation for larger systems. Additionally, their simplistic action settings focus on vulnerability exploitation without addressing the logical relationships among multiple penetration tactics.

\begin{figure*}[tb]
    \centering
    \includegraphics[width=\linewidth]{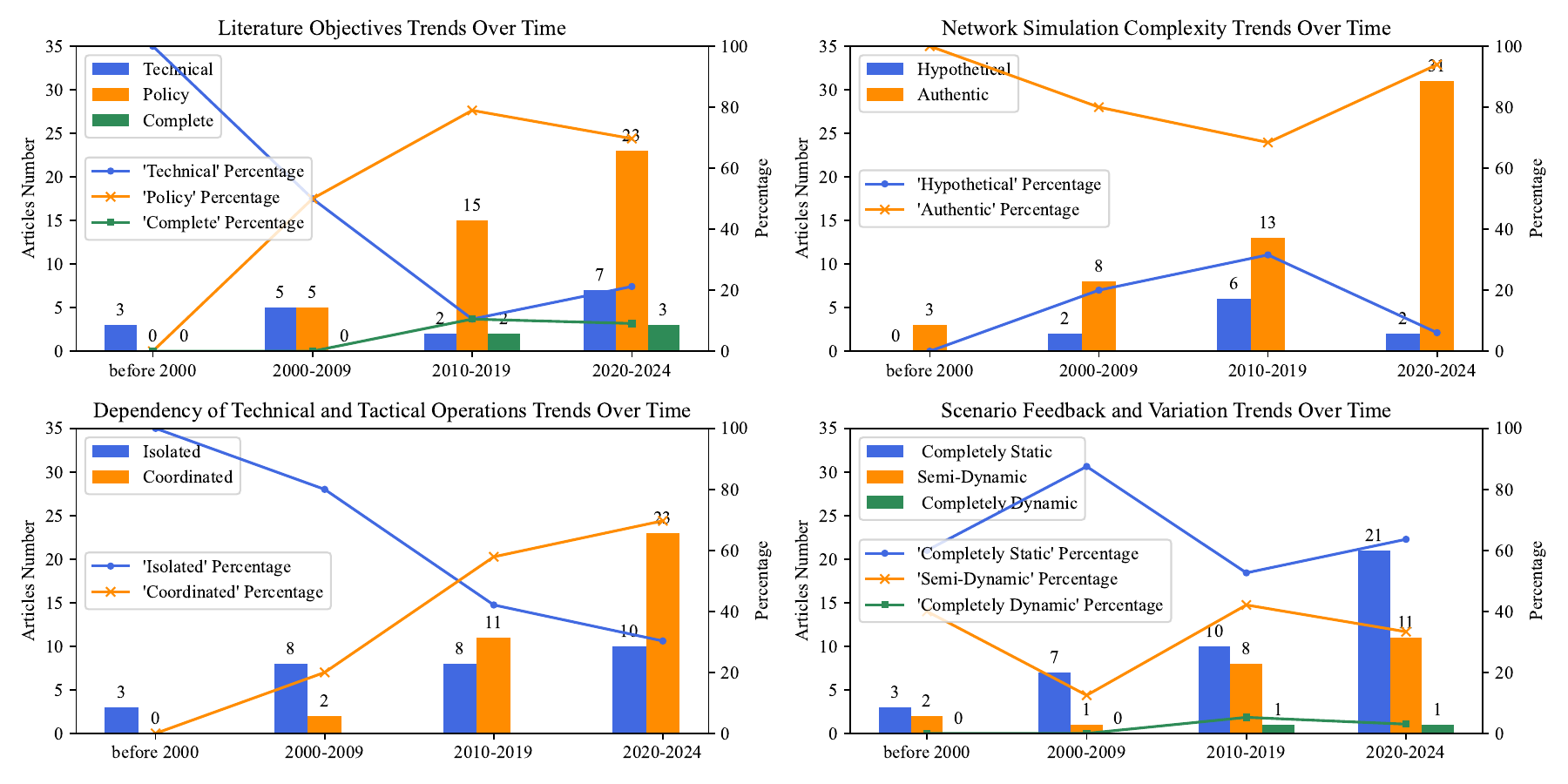}
    \caption{Temporal Variations in Article Volume Across Dimensions}
    \label{fig: Dimensions Over Time}
\end{figure*}

Table~\ref{Classification Statistics Table} summarizes literature classifications over time, while Figure~\ref{fig: Dimensions Over Time} shows trends in article volumes across dimensions. Initially, AutoPT research focused on Technical Automation, emphasizing the automation of specific tactics and steps. Over time, Policy Automation became more prominent, evolving from simulating hypothetical attributes and isolated actions to simulating authentic attributes and coordinated technical and tactical actions. This shift reflects reduced abstraction in simulation modeling and a closer alignment between models and reality, paving the way for integrating intelligent decision-making algorithms with automated tools for Complete Automation. Ongoing research in Complete Automation typically streamlines the AutoPT process by targeting specific components, incorporating one or more automated tools—such as rules, planners, or basic reinforcement learning models—to develop action-guiding strategies with a greater emphasis on engineering implementation and lower intelligence levels. Notably, Ghanem et al.~\cite{ghanem2019reinforcement} focus on automating strategy generation using tools like MSF for execution. Although Complete Automation is not fully achieved, their work is practically significant and classified under Complete Automation, demonstrating the flexibility of our criteria.


Besides, most studies focus on static or semi-dynamic scenarios, neglecting active network changes and dynamic information. Approaches like Cyber Mimic Defense~\cite{2016Research} and Moving Target Defense~\cite{2011Moving} signal a shift toward dynamic network security strategies. Future research should integrate both active and passive network changes to enable intelligent decision-making and automated execution in fully dynamic environments. Additionally, existing studies typically simulate only small to medium-sized networks~\cite{Cyberbattlesim, li2024dynpen, applebaum2017analysis}, overlooking large-scale network modeling and the impact of diverse network architectures on penetration testing. Furthermore, current simulation methods lack flexibility, focusing on limited combinations without providing a unified approach for multi-dimensional and multi-level simulation modeling.





\section{AutoPT-Sim: A Unified Simulation Modeling Framework for Automated Penetration Testing}

\begin{figure*}[tb]
    \centering
    \includegraphics[width=150mm]{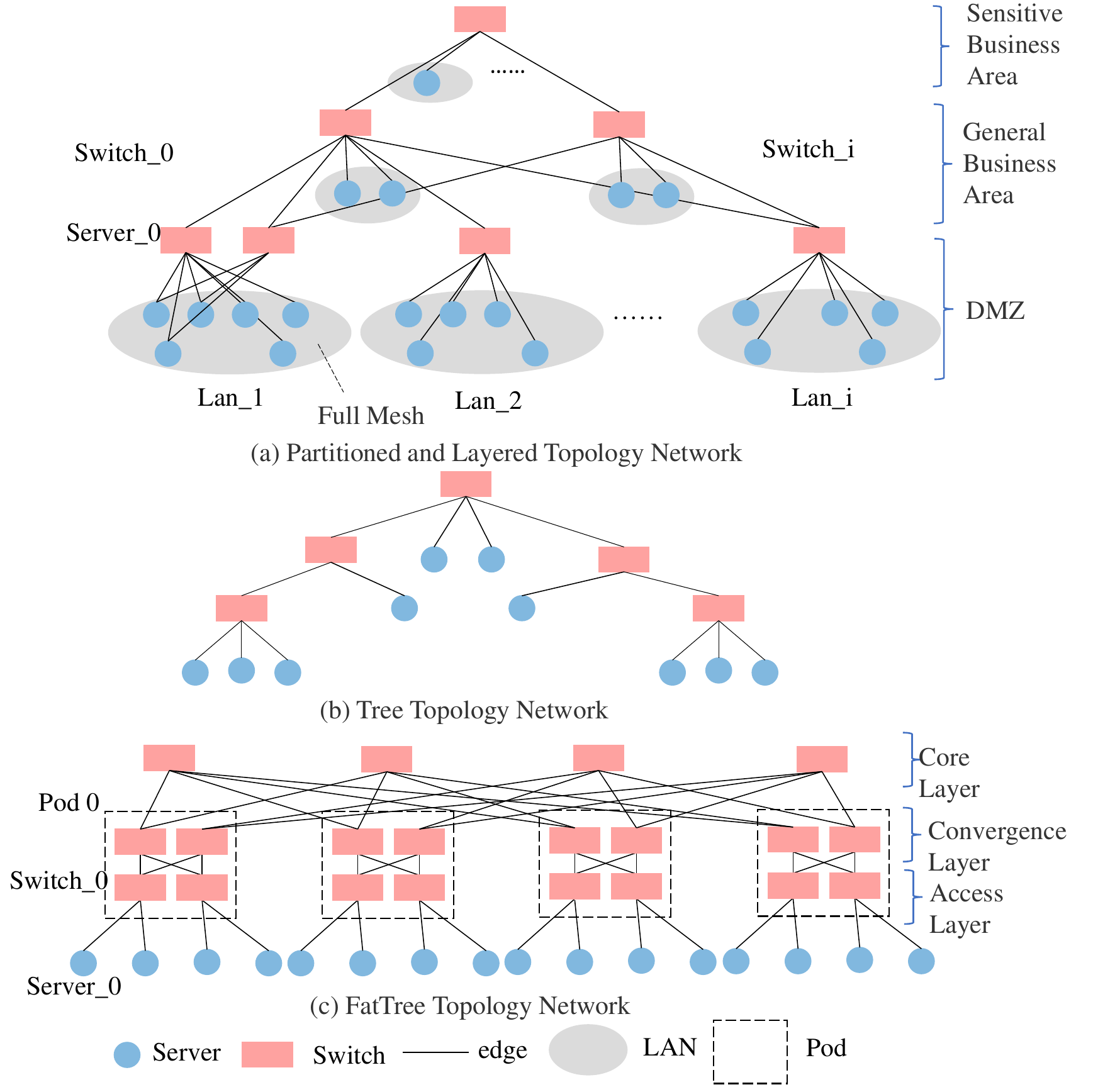}
    \caption{Three Types of Network Topology in~\modelsim}
    \label{three_kind_topology}
\end{figure*}

Current scenario modeling methods are often incomplete, exhibit low variability, lack fully dynamic capabilities, and are hindered by the absence of public datasets. To address these limitations, we introduce the AutoPT Simulation Modeling Framework (\modelsim), which leverages policy automation to integrate all sub-dimensions of the \modelcla~framework across its three primary dimensions. Additionally, we provide a comprehensive and publicly accessible dataset on GitHub\footnote{www.github.com/feifei-feifei-hub/Simulation-Modeling-for-Automated-Penetration-Testing} to support future research endeavors. We welcome constructive feedback to refine our standards and dataset further. Furthermore, we have developed a suite of interfaces to underpin future extensions in tactical and full-process automation.



\subsection{AutoPT Simulation Modeling Framework}
Our model enables AutoPT research within internal networks by automating policy scenarios and simulating all penetration testing phases, including information gathering, foothold establishment, privilege escalation, lateral movement, and persistence. We provide diverse network architectures and asset modeling techniques to support dynamic network construction. Additionally, attacker and defender actions are comprehensively modeled, allowing for customizable configurations.




\subsubsection{Network Architecture and Target Asset Modeling}
We model the target network by combining its architecture and assets into a graph $G = (V, E, X)$, where nodes $V$ represent network devices, edges $E$ denote their connections, and $X$ captures attributes for both nodes and edges. Computer networks naturally form graph structures~\cite{2024Graph}. Each node $V = \{1, 2, 3, \dots, N\}$ corresponds to a device, with attributes $X_i$ detailing systems, services, and credentials. Edges $E$ represent both wired and wireless communications. An edge $e_{ij}$ exists if node $i$ can communicate with node $j$, with link attributes $X_{ij}$ specifying protocols, traffic size, and more. All nodes have a type attribute distinguishing categories like server or switch, effectively capturing configurations and functionalities through attributes such as software and services.


To emulate attackers' extensive maneuverability in internal networks, all connections are bidirectional. Our penetration testing framework emphasizes lateral movement by focusing on node configurations and attributes rather than simulating traffic. We incorporate various network topologies to reflect real-world diversity, and our network generator allows for the expansion of directed links and link attributes to represent data flow characteristics. The specific methods for network architecture and target asset modeling are detailed below.



\textbf{Network Architecture Modeling.}
Our simulated network model assumes direct connections between nodes on the same LAN and switch-mediated communication between nodes on different LANs. The network graph features undirected edges, enabling bidirectional communication between nodes. To capture the diversity of real-world network topologies, our framework incorporates multiple classic topologies, as well as partitioned and layered topologies informed by expert insights. These topologies are illustrated in Figure~\ref{three_kind_topology}.

\begin{itemize}
    \item \textbf{Partitioned and Layered Topology Network}: We employ a customized network topology, initially introduced by Sarraute et al.~\cite{sarraute2012pomdps} and subsequently refined through expert consultation with cybersecurity professionals. The PLTN architecture is specifically designed for performance testing and comprises three distinct regions: (1) the Demilitarized Zone (DMZ), (2) the General Business Area, and (3) the Sensitive Business Area.

        \begin{enumerate}
            \item \textbf{DMZ} connects the external internet to the general business area. It features interconnected nodes with minimal defenses, lenient firewall rules, and lower account privileges, making it a potential entry point but less likely to contain sensitive information.
            \item \textbf{General Business Area} acts as a bridge between the DMZ and the sensitive business area. It includes multiple subnets with enhanced defenses, some sensitive data, and high-level accounts. Connections between subnets are managed by devices like firewalls and routers.
            \item \textbf{Sensitive Business Area} connects only to the general business area and consists of 1-3 subnets with the strongest defenses and strict access controls. It has fewer nodes but is more likely to contain sensitive information, with connectivity managed similarly to the general business area.
        \end{enumerate}
        Backup switches enhance network robustness by demonstrating variability in node connections across different layers and LANs. This setup reflects the network's inherent randomness and adaptability.

    \item \textbf{Tree Topology Network}~\cite{Brede2012NetworksAnIM}: A classic network architecture where nodes are arranged in layers, typically consisting of a root, branch, and leaf nodes. Each node has a unique data transmission path, simplifying traffic control and management. The failure of a node or link affects only its subtree. Common in broadband networks like Ethernet, the central node is usually a switch or hub, with branches and leaves as workstations or computers. While easy to manage, a central node failure can jeopardize the entire network.


    \item \textbf{FatTree Topology Network}~\cite{al2008scalable}: This scalable data center network architecture addresses traditional topology limitations, offering improved scale and bandwidth. By employing multiple low-cost units, it builds a large-scale structure ideal for high-performance computing and big data tasks. The FatTree is a $k$-ary tree with $k$ ports per switch, $(k/2)^2$ core switches, and $k$ pods. Each pod contains two layers: the aggregation layer and the access layer (or edge layer), each with $k/2$ switches. Aggregation layer ports connect to core switches and access layer switches. In the access layer, ports connect to the aggregation layer and hosts. This design enhances network performance with scalable bandwidth and hierarchical connectivity.

\end{itemize}

We use the topology type as input for the network generator, enabling researchers to easily create diverse network topologies and utilize implicit information within them.

\textbf{Target Asset Modeling.}
Our framework models nodes in a network graph using hypothetical and authentic attribute simulations.

\textbf{Simulation of Hypothetical Attributes}: We follow Hammar et al.~\cite{2020Finding} to assign node attributes, creating networks with hypothetical attributes. Each node is assigned $m+1$ values: the first $m$ reflect its defensive capabilities, and the last indicates its anomaly detection capability. If a node has a vulnerability in the $i$-th defense, the $i$-th attribute value is set to $x_i \leq 1$. 

\textbf{Simulation of Authentic Attributes}: Leveraging expert insights, we identify critical attributes for penetration testing, setting unique attributes for each node to simulate authentic target assets:

\begin{itemize}
    \item \textbf{IP} ($ip$): The IP address of the node.
    \item \textbf{Node Type} ($type$): Nodes are classified as either switch for data transmission or server for processing and storage. Customizable types are supported.
    \item \textbf{Local Area Network ID} ($lan\_id$): Identifies the node's LAN, reflecting our partitioned network design. Nodes in the same LAN are assumed connected, despite possible internal firewalls.
    \item \textbf{Operating System} ($system$): Categorized as $windows$, $linux$, or $other$.
    \item \textbf{Open Ports and Services} ($port\_server\_version$): Lists the node’s open ports and associated services, including service versions. Scans may omit this information, particularly versions, so we simulate potential data loss with a predefined probability.
    \item \textbf{Installed Software and Versions} ($software\_version$): Details additional software on the node that doesn't provide external services and their versions, which might have vulnerabilities or sensitive data like passwords, accessible through specific methods.
    \item \textbf{Saved Account Passwords and Levels} ($account$): Encompasses standard, administrative, and domain administrator accounts.
     \item \textbf{Vulnerabilities and Vulnerability Exploit Success Rate} ($cve$): Indicates vulnerabilities related to the node's OS, services, software and weak passwords, alongside the likelihood of successful exploitation. We collected more than 500 vulnerabilities across various systems and services. CVSS scores assess severity based on factors like attack vector and complexity, while EPSS scores, from the latest EPSS v3 model, estimate exploitation likelihood in the wild, with higher scores indicating greater risk. The EPSS score from 2024.10, and CVSS  scores are used to determine exploitation success probability.
\end{itemize}

For example, the attributes $ X_i $ can be set as follows:
\begin{equation}
    \begin{aligned}
        X_i = \{
        'ip'&: '192.168.2.105',\\
        'type'&: 'server', \\
        'lan\_id'&: '9', \\
        'system'&: 'linux',\\
        'port\_server\_version'&: [('3306', 'Cacti', '1.2.22'),\\
        & ('6699', 'samba', '3.5.0') ],\\
        & ('22', 'ssh', 'Tom', 'Tom123') ],\\
        'software\_version'&:[('Struts2', '2.0.0'), \\
        &('PostgreSQL', '9.3')],\\
        'account'&:[('Tom','123QWE','root')],\\
        'cve'&:[('CVE-2022-46169',0.65), \\
        &('CVE-2017-7494',0.80)]\}
    \end{aligned}
\end{equation}

To enhance network attribute generation, we incorporate an underlying pattern based on expert knowledge. Nodes within the same LAN often exhibit similarities, particularly concerning operating systems and installed software, such as Windows and Office. This suggests they may share common system and software vulnerabilities, mirroring real-world scenarios where nodes in the same department have similarities. However, due to varying user habits, some vulnerabilities may be addressed through system patches or software updates, and this aspect is also simulated in our network model.
Link attributes can be configured based on research needs and decision-making methods. The selection of node and link attributes for intelligent decision-making should align with the specific method and scenario.


\subsubsection{Attacker and Defender Modeling}




In this section, we model attacker and defender actions by integrating established methods from literature and expert insights.
For hypothetical attribute networks, we follow the settings by Hammar et al.~\cite{2020Finding}. Attackers increment a node's attack value by 1 in a specific dimension. An attack succeeds when this value exceeds the node's defense value. Defenders can add a defense value of 1 to a node or detect attackers probabilistically. For real attribute networks, we employ a more complex and realistic modeling approach for attackers.



\textbf{Attacker Modeling.}
Our attacker modeling is based on the MITRE ATT\&CK framework, which outlines 14 tactic and technique phases.
Below are the methods applicable at each phase, the necessary execution details, and the results obtained.

\begin{table*}[t]
\centering
\caption{Attacker Action Modeling}
\label{tab:attacker_action_modeling}
\resizebox{\linewidth}{!}{%
\begin{tabular}{p{1.5cm}|p{3cm}|p{2cm}|p{1.4cm}|p{2.5cm}|p{2cm}|p{0.5cm}|p{1.2cm}|p{3.5cm}}
\toprule[2pt]
\textbf{Action} & \textbf{ATT\&CK Phase} & \textbf{Precondition} & \textbf{Decision Parameters} & \textbf{Expected Outcomes} & \textbf{Unexpected Outcomes} & \textbf{Time Cost} & \textbf{Network Changed} & \textbf{Note} \\ \hline
Scanning & Reconnaissance, Discovery & Target IP known & Target IP & Obtain partial information of the target IP,including type, LAN ID, system, and (port, service, version) attributes. &  & 2 & No & Version data is often returned probabilistically, simulating randomness and potential pairing errors. Similarly, exposure surface information is provided in a probabilistic manner, mirroring the challenges of incomplete data in real-world scanning tools.
 \\ \hline
Vulnerability Exploitation & Initial Access, Execution, Persistence, Privilege Escalation, Lateral Movement, Command and Control & Target IP known & Target IP, specific vulnerability & Obtain control permissions of the target IP node & If the vulnerability does not exist, exploitation fails & 1 & No & Success judged by CVSS and EPSS scores; user-set success probability supported. \\ \hline
Persistence & Persistence & Administrative user permissions for target IP & Target IP & Obtain persist session & Persistent session not retrieved & 1 & No & Node maintains session post-restart. \\ \hline
Credential Access & Initial Access, Privilege Escalation, Credential Access, Lateral Movement & Target IP known & Target IP & Obtain different permissions based on (account, password) level & Login fails if credentials do not match & 3 & No & Success based on target IP's credentials in repository. \\ \hline
Weak Password Cracking & Initial Access, Privilege Escalation, Lateral Movement & Target IP known & Target IP & Gain administrative privileges & Login fails if password does not match weak password database & 3 & No & Success based on weak password vulnerability. \\ \hline
Obtain Sensitive Information & Collection, Command and Control, Discovery & Permissions for target IP & Target IP & Obtain all attributes and sensitive information; add host credentials to database & & 2 & No & Support partial information return to demonstrate attacker capabilities. \\ \hline
Phishing Email & Privilege Escalation, Lateral Movement, Initial Access & Target IP known & Target IP & Obtain different privileges & & 2 & No & Success judged by target IP node attributes and success rate. \\ \hline
Information Leakage & Command and Control, Exfiltration, Impact & Sensitive information for target IP & Target IP & Leak sensitive information & & 2 & No & Based on researcher's objectives. \\ \hline
Establish or Disconnect Connection & Command and Control, Lateral Movement & Permissions for target IP & Initial IP, Target IP & Create or disconnect connections & &1& Yes & \\ \hline
Force Host Offline/online & Command and Control, Impact & Permissions for target IP & Target IP & Target node goes offline or online & & 1 & Yes & Causes network paralysis, affecting services. \\ \hline
Defense Evasion & Defense Evasion & Permissions for target IP & Target IP & Clean up action traces to reduce discovery probability & & 2 & No & \\ \bottomrule[2pt]
\end{tabular}
}
\end{table*}

\begin{itemize}
    \item \textbf{Reconnaissance:} Input the target node IP and return its attack surface attributes. Ports and services are paired, and service software versions have a 50\% chance of being returned.
    \item \textbf{Resource Development:} Establish resources like fingerprint vulnerabilities, weak password libraries, and payloads collected before penetration testing. This is a preparatory phase rather than a decision-making stage in intelligent decision processes. During penetration, experts choose suitable resources and tools from what's available. The thoroughness of this preparation dictates the range of potential actions.
    \item \textbf{Initial Access:} Achieved via actions such as exploiting vulnerabilities, phishing, cracking weak passwords, and using credentials.
    \item \textbf{Execution:} Indicates that the attacker has gained initial access to a target network node and is running attack-related code. This is typically accomplished by exploiting vulnerabilities to subsequently gather system information, sensitive data, and additional node content.
    \item \textbf{Persistence:} Determine which nodes to maintain access to for ongoing exploitation.
    \item \textbf{Privilege Escalation:} Decide where to escalate privileges using vulnerabilities, password cracking, or credential login.
    \item \textbf{Defense Evasion:} Erase traces of actions on a node to reduce detection chances.
    \item \textbf{Credential Access:} Use credentials such as passwords, cookies, tokens, tickets, and cryptographic elements including hashes, keys, certificates, fingerprints, and biometric data to obtain node access permissions. In modeling, we use the 'Credential Access' action to encompass these various attack methods.
    \item \textbf{Discovery:} Internal network reconnaissance by gathering system and network info through techniques like discovery of account, address space, URL, and system, aiding in mapping environments of compromised hosts. 
    \item \textbf{Lateral Movement:} Select a host to move to, decide on the target node and the access method..
    \item \textbf{Collection:} Attackers collect valuable information, such as drive types, browsers, audio, video, email, and file contents. In modeling, we use "Obtain Sensitive Information" action to encompass all techniques at this stage and restrict the information gathered to the internal data of the controlled host, distinguishing it from the externally exposed surface data collected via active scanning during the Renaissance or Discovery phases.
    \item \textbf{Command and Control:} Remotely control the host to execute commands and operations, potentially leading to information leakage and connection manipulation by exploiting vulnerabilities.
    \item \textbf{Exfiltration:} Decide on transmitting sensitive information externally after collection.
    \item \textbf{Impact:} Aim to manipulate, disrupt, or interfere with systems and data.
\end{itemize}


Based on our analysis of intelligent decision-making needs, decision content, methods, common attack actions in research, and real-life penetration parameters, we provide a detailed list of attacker actions in Table~\ref{tab:attacker_action_modeling}.





\textbf{Defender Modeling.}
Defender modeling integrates research and practical defense strategies.
Network administrators enhance detection and defense by patching vulnerabilities, deploying intrusion detection systems (IDS), and monitoring traffic. They terminate attacker sessions by taking nodes offline, blocking IPs, and clearing login credentials to protect compromised nodes. Detailed actions are in Table~\ref{tab:defender_actions}.

\begin{table*}
\centering
\caption{Defender Action Modeling}
\label{tab:defender_actions}
\resizebox{\linewidth}{!}{%
\begin{tabular}{p{2.5cm}|p{2.5cm}|p{4cm}|p{1cm}|p{1cm}|p{6cm}} 
\toprule[1.5pt]
Action &  Decision Parameters & Expected Outcomes & Time Cost& Network Changed & Note \\ \midrule[1.2pt]
Patch Vulnerability &  Target IP & Patch a specific vulnerability &1 & Yes & One vulnerability of the target IP is randomly patched at a time. This invalidates sessions established through that vulnerability, causing the attacker to lose control of the host. \\ \hline
Traffic Monitoring &  Target IP & Monitor node traffic and invalidate suspicious sessions & 2 &No & Attackers may cause suspicious traffic changes when using information leakage. \\ \hline
Detect Attack &  Target IP & If the detected target IP is n=2 time steps behind the attack's vulnerability exploitation, the foothold is invalidated &1 & No & \\ \hline
Proactively Take Host Offline &  Target IP & Take a specific host offline & 1 &Yes & Target IP will come back online after a five time steps interval. Upon reconnection, the attacker will lose control of the host, and all established sessions will be terminated.\\ \hline
IP Blacklisting  &  Target IP & Randomly disconnect one connection of the target IP node & 1 &Yes & \\ \hline
Clear/Add Active Credentials &  Target IP, Clear/Add Active Credentials & Clear/Add active credentials for the target IP & 1 &Yes & \\ \hline
Honeypot&  Target IP& Configure the target IP as a honeypot; an alarm message will trigger upon a successful attacker penetration of the node.& 1 & No &One of the conditions for penetration failure can be defined as an attacker successfully infiltrating a honeypot.\\ \hline
Countermeasure&Target IP& Obtain the attacker's IP and related information & 2 & No &Upon honeypot infiltration, the defender implements countermeasures to pinpoint the attacker's IP address andrelated information, which can be designed as an ending condition.\\ \hline
Network security training& None& Randomly reduce the success rate of attack methods such as phishing emails, weak passwords, and credential login. & 10&No & The degree of success rate reduction varies for each node.\\ \bottomrule[1.5pt]
\end{tabular}}
\end{table*}


In real-world scenarios, the network is visible to defenders, but attackers' actions are concealed. Traditional AutoPT research models defender awareness through detection. Our approach enhances defense capabilities and detection, incorporating proactive measures like IP blacklisting, honeypots, and countermeasures to capture attacker traces. We have also integrated social engineering defense by providing extensive security education to reduce phishing success rates and strengthen network security.


In addition to measures shown in Table~\ref{tab:defender_actions}, some implicit defenses can be strengthened by defining communication relationships between nodes. Dynamic target defense and network mimicry can be implemented through adaptive network changes. In zero trust environments, this contemporary approach enhances defense capabilities via continuous and dynamic verification processes. Our framework simulates continuous authentication by regularly updating node credentials, limiting attackers to credential-logging for session persistence. By adjusting inter-node communication and simulating granular access controls, we can effectively model zero trust scenarios.


Certain attacker and defender actions can passively modify the network architecture and target assets, as shown in Tables \ref{tab:attacker_action_modeling} and \ref{tab:defender_actions}. During modeling, not all actions are essential for a complete penetration phase. It is preferable to select actions relevant to the specific application phase and research method. Simplifying decision parameters enhances the effective use of independent and continuous tactics and techniques.

\subsection{Network Simulation Dataset and Network Generator}

We developed a network generator using the \modelsim~framework, enabling the creation of dynamic and static networks with diverse architectures, attributes, and scales. The open-source code allows researchers to generate custom network data through parameter adjustments, thereby advancing AutoPT research. We offer a pre-generated network simulation dataset, which includes hypothetical numerical attributes, authentic attributes, and their continuous-time counterparts.
\begin{itemize}
    \item Static Hypothetical Numerical Attributes Simulation Networks: These are based on numerical simulations with hypothetical attributes, with no active changes in network scenarios. 
    \item Static Authentic Attributes Simulation Networks: These use authentic attributes without active changes in network scenarios.
    \item Dynamic Hypothetical Numerical Simulation Networks: These incorporate hypothetical attributes alongside dynamic scenario alterations. Nodes may be added or modified randomly, affecting connections and attributes according to a specified change proportion, $p_{\text{change}}$.
    \item Dynamic Authentic Attribute Simulation Networks: Here, authentic attributes are used with dynamic scenario changes, governed by $p_{\text{change}}$.
\end{itemize}
Based on the aforementioned configurations, we include three types of networks with scales of 10, 100, and 1,000 nodes in our dataset. 
These scales can also be expanded by modifying the scale parameter. 
For dynamic networks, we produce network graphs at various time points, providing snapshots that represent the network's evolution. Researchers can switch between these snapshots to effectively capture network dynamics. 




\textbf{Usage Example.}
To construct a simulation scenario with policy automation, authentic attributes, coordinated technical and tactical actions, and a semi-dynamic context, start by using datasets from static authentic attribute networks. Then, select actions from the attacker and defender sets while ensuring they meet preconditions, and incorporate at least one action to induce passive changes in the network structure.

To advance simulation modeling in policy automation within~\modelcla, we integrate publicly available datasets flexibly. We have also released network generator code, enabling researchers to customize network data by adjusting parameters or fine-tuning the generator. \modelsim~addresses the limitations of existing scene modeling methods, which often focus on small to medium-sized networks and lack support for large networks and varied architectures. Current methods fall short in dynamic scene modeling, offering limited flexibility and lacking a unified approach for multidimensional and multi-layer simulations.

\section{Conclusion}

This paper reviews the literature on AutoPT and introduces an innovative classification framework, \modelcla, for scenario modeling methods. Our framework categorizes existing research distinctly, addressing the limited scope, fragmented scenarios, and lack of unified standards and public datasets in current AutoPT modeling. We propose~\modelsim, a method that emphasizes strategy automation while supporting tactic and technique automation, as well as full-process integration. Our public release includes a network scenario dataset and network generator code, facilitating flexible scenario modeling across all levels and enabling researchers to customize network data by adjusting generator parameters.
Our construction method and dataset aim to guide simulation modeling in AutoPT and serve as a standard data benchmark for fair comparisons of intelligent decision-making methods. To our knowledge, this is the first work to analyze and classify simulation modeling in AutoPT, while offering guidance and standard datasets for model construction.

Ethical considerations are crucial in AutoPT. Our modeling framework, \modelsim, uses real-world data while abstracting it to protect privacy and excludes actual penetration tools and payloads, ensuring no direct real-world application. Our research focuses on developing penetration strategies without full automation, thereby avoiding potential harm to systems or users.

In our current modeling framework, we detail the modeling of attacker and defender actions and plan to release the attacker-defender action dataset and state transition functions in a future phase. We will enhance our characterization of attacker capabilities post-intrusion. Recognizing that information gathering significantly depends on the attacker's expertise, we will examine network visibility disparities between attackers and defenders, as well as defenders' delayed response times to attacker actions.
Another significant challenge in AutoPT is the lack of a unified evaluation method. Current evaluations often emphasize convergence speed and cumulative rewards in specific network settings, with varying reward configurations complicating comparisons. There is no widely accepted set of metrics for assessing the effectiveness of intelligent decision-making. We advocate for future research to develop standardized evaluation metrics to enhance comparison and advance AutoPT methodologies.





\bibliographystyle{ieeetr}
\bibliography{reference}

\begin{thebibliography}{10}

\bibitem{chen2022research}
Z.~Chen, ``Research on internet security situation awareness prediction technology based on improved rbf neural network algorithm,'' {\em Journal of Computational and Cognitive Engineering}, vol.~1, no.~3, pp.~103--108, 2022.

\bibitem{wani2021sdn}
A.~Wani, R.~S, and R.~Khaliq, ``Sdn-based intrusion detection system for iot using deep learning classifier (idsiot-sdl),'' {\em CAAI Transactions on Intelligence Technology}, vol.~6, no.~3, pp.~281--290, 2021.

\bibitem{verma2024revisiting}
R.~Verma, A.~Kumari, A.~Anand, and V.~Yadavalli, ``Revisiting shift cipher technique for amplified data security,'' {\em Journal of Computational and Cognitive Engineering}, vol.~3, no.~1, pp.~8--14, 2024.

\bibitem{abu2018automated}
F.~Abu-Dabaseh and E.~Alshammari, ``Automated penetration testing: An overview,'' in {\em The 4th international conference on natural language computing, Copenhagen, Denmark}, pp.~121--129, 2018.

\bibitem{dorchuck2021goal}
S.~J. Dorchuck, {\em Goal-Directed Systems Testing: Automated Execution of Intelligently Generated Cyber Attack Plans}.
\newblock PhD thesis, Massachusetts Institute of Technology, 2021.

\bibitem{applebaum2016intelligent}
A.~Applebaum, D.~Miller, B.~Strom, C.~Korban, and R.~Wolf, ``Intelligent, automated red team emulation,'' in {\em Proceedings of the 32nd Annual Conference on Computer Security Applications}, pp.~363--373, 2016.

\bibitem{ghanem2018reinforcement}
M.~C. Ghanem and T.~M. Chen, ``Reinforcement learning for intelligent penetration testing,'' in {\em 2018 second world conference on smart trends in systems, security and sustainability (WorldS4)}, pp.~185--192, IEEE, 2018.

\bibitem{hacks2021towards}
S.~Hacks, R.~Lagerstr{\""o}m, and D.~Ritter, ``Towards automated attack simulations of bpmn-based processes,'' in {\em 2021 IEEE 25th International Enterprise Distributed Object Computing Conference (EDOC)}, pp.~182--191, IEEE, 2021.

\bibitem{applebaum2017analysis}
A.~Applebaum, D.~Miller, B.~Strom, H.~Foster, and C.~Thomas, ``Analysis of automated adversary emulation techniques,'' in {\em Proceedings of the summer simulation multi-conference}, pp.~1--12, 2017.

\bibitem{yichao2019improved}
Z.~Yichao, Z.~Tianyang, G.~Xiaoyue, and W.~Qingxian, ``An improved attack path discovery algorithm through compact graph planning,'' {\em IEEE Access}, vol.~7, pp.~59346--59356, 2019.

\bibitem{obes2013attack}
J.~L. Obes, C.~Sarraute, and G.~Richarte, ``Attack planning in the real world,'' {\em arXiv preprint arXiv:1306.4044}, 2013.

\bibitem{hu2020automated}
Z.~Hu, R.~Beuran, and Y.~Tan, ``Automated penetration testing using deep reinforcement learning,'' in {\em 2020 IEEE European Symposium on Security and Privacy Workshops (EuroS\&PW)}, pp.~2--10, IEEE, 2020.

\bibitem{chowdhary2020autonomous}
A.~Chowdhary, D.~Huang, J.~S. Mahendran, D.~Romo, Y.~Deng, and A.~Sabur, ``Autonomous security analysis and penetration testing,'' in {\em 2020 16th International Conference on Mobility, Sensing and Networking (MSN)}, pp.~508--515, IEEE, 2020.

\bibitem{bianou2024pentest}
S.~G. Bianou and R.~G. Batogna, ``Pentest-ai, an llm-powered multi-agents framework for penetration testing automation leveraging mitre attack,'' in {\em 2024 IEEE International Conference on Cyber Security and Resilience (CSR)}, pp.~763--770, IEEE, 2024.

\bibitem{deng2024pentestgpt}
G.~Deng, Y.~Liu, V.~Mayoral-Vilches, P.~Liu, Y.~Li, Y.~Xu, T.~Zhang, Y.~Liu, M.~Pinzger, and S.~Rass, ``Pentestgpt: Evaluating and harnessing large language models for automated penetration testing,'' in {\em 33rd USENIX Security Symposium (USENIX Security 24)}, pp.~847--864, 2024.

\bibitem{chenke2023survey}
k.~Chen, H.~Lu, B.~Fang, Y.~Sun, s.~Su, and Z.~Tian, ``Survey on automated penetration testing technology research,'' {\em Journal of Software}, pp.~1--21, 2023.

\bibitem{ErikMiehling2019Control}
ErikMiehling, MohammadRasouli, DemosthenisTeneketzis, ErikMiehling, MohammadRasouli, DemosthenisTeneketzis, ErikMiehling, MohammadRasouli, DemosthenisTeneketzis, and E.~and, ``Control-theoretic approaches to cyber-security,'' 2019.

\bibitem{furfaro2017using}
A.~Furfaro, L.~Argento, A.~Parise, and A.~Piccolo, ``Using virtual environments for the assessment of cybersecurity issues in iot scenarios,'' {\em Simulation Modelling Practice and Theory}, vol.~73, pp.~43--54, 2017.

\bibitem{miller2018automated}
D.~Miller, R.~Alford, A.~Applebaum, H.~Foster, C.~Little, and B.~Strom, ``Automated adversary emulation: A case for planning and acting with unknowns,'' {\em MITRE CORP MCLEAN VA MCLEAN}, 2018.

\bibitem{2020Finding}
K.~Hammar and R.~Stadler, ``Finding effective security strategies through reinforcement learning and self-play,'' 2020.

\bibitem{Brede2012NetworksAnIM}
M.~Brede, ``Networks—an introduction. mark e. j. newman. (2010, oxford university press.) \$65.38, £35.96 (hardcover), 772 pages. isbn-978-0-19-920665-0.,'' {\em Artificial Life}, vol.~18, pp.~241--242, 2012.

\bibitem{Guo2018Cyberspace}
L.~Guo, Y.~Cao, M.~Su, Y.~Shang, Y.~Zhu, P.~Zhang, and C.~Zhou, ``Cyberspace resource mapping: Concepts and techniques,'' {\em Journal of Information Security}, vol.~3, no.~4, p.~14, 2018.

\bibitem{2016Research}
W.~U. Jiangxing, ``Research on cyber mimic defense,'' {\em Journal of Cyber Security}, 2016.

\bibitem{2011Moving}
S.~Jajodia, A.~K. Ghosh, V.~Swarup, C.~Wang, and X.~S. Wang, {\em Moving Target Defense: Creating Asymmetric Uncertainty for Cyber Threats}.
\newblock Moving Target Defense: Creating Asymmetric Uncertainty for Cyber Threats, 2011.

\bibitem{hutchins2011intelligence}
E.~M. Hutchins, M.~J. Cloppert, R.~M. Amin, {\em et~al.}, ``Intelligence-driven computer network defense informed by analysis of adversary campaigns and intrusion kill chains,'' {\em Leading Issues in Information Warfare \& Security Research}, vol.~1, no.~1, p.~80, 2011.

\bibitem{2018MITRE}
B.~E. Strom, A.~Applebaum, D.~Miller, K.~C. Nickels, A.~G. Pennington, and C.~Thomas, ``Mitre att\&ck : Design and philosophy,'' 2018.

\bibitem{Midian}
P.~Midian, ``Perspectives on penetration testing — black box vs. white box.,'' {\em Network Security}, p.~10, 2002.

\bibitem{al2018study}
H.~M.~Z. Al~Shebli and B.~D. Beheshti, ``A study on penetration testing process and tools,'' in {\em 2018 IEEE Long Island Systems, Applications and Technology Conference (LISAT)}, pp.~1--7, IEEE, 2018.

\bibitem{filiol2021method}
E.~Filiol, F.~Mercaldo, and A.~Santone, ``A method for automatic penetration testing and mitigation: A red hat approach,'' {\em Procedia Computer Science}, vol.~192, pp.~2039--2046, 2021.

\bibitem{shravan2014penetration}
K.~Shravan, B.~Neha, and B.~Pawan, ``Penetration testing: A review,'' {\em Compusoft}, vol.~3, no.~4, p.~752, 2014.

\bibitem{goel2015vulnerability}
J.~N. Goel and B.~M. Mehtre, ``Vulnerability assessment \& penetration testing as a cyber defence technology,'' {\em Procedia Computer Science}, vol.~57, pp.~710--715, 2015.

\bibitem{demott2007revolutionizing}
J.~DeMott, R.~Enbody, and W.~F. Punch, ``Revolutionizing the field of grey-box attack surface testing with evolutionary fuzzing,'' {\em BlackHat and Defcon}, 2007.

\bibitem{awang2013detecting}
N.~F. Awang and A.~A. Manaf, ``Detecting vulnerabilities in web applications using automated black box and manual penetration testing,'' in {\em International Conference on Security of Information and Communication Networks}, pp.~230--239, Springer, 2013.

\bibitem{terranova2024leveraging}
F.~Terranova, A.~Lahmadi, and I.~Chrisment, ``Leveraging deep reinforcement learning for cyber-attack paths prediction: Formulation, generalization, and evaluation,'' in {\em The 27th International Symposium on Research in Attacks, Intrusions and Defenses (RAID 2024)}, 2024.

\bibitem{ghanem2022towards}
M.~C. Ghanem, {\em Towards an efficient automation of network penetration testing using model-based reinforcement learning}.
\newblock PhD thesis, City, University of London, 2022.

\bibitem{dillon2022perihack}
R.~Dillon {\em et~al.}, ``“perihack”: Designing a serious game for cybersecurity awareness,'' in {\em 2022 IEEE International Conference on Teaching, Assessment and Learning for Engineering (TALE)}, pp.~630--634, IEEE, 2022.

\bibitem{schwartz2019autonomous}
J.~Schwartz and H.~Kurniawati, ``Autonomous penetration testing using reinforcement learning,'' {\em arXiv preprint arXiv:1905.05965}, 2019.

\bibitem{paul2019learning}
S.~Paul, Z.~Ni, and C.~Mu, ``A learning-based solution for an adversarial repeated game in cyber--physical power systems,'' {\em IEEE Transactions on Neural Networks and Learning Systems}, vol.~31, no.~11, pp.~4512--4523, 2019.

\bibitem{elderman2017adversarial}
R.~Elderman, L.~J. Pater, A.~S. Thie, M.~M. Drugan, and M.~A. Wiering, ``Adversarial reinforcement learning in a cyber security simulation,'' in {\em 9th International Conference on Agents and Artificial Intelligence (ICAART 2017)}, pp.~559--566, SciTePress Digital Library, 2017.

\bibitem{lyon2009nmap}
G.~F. Lyon, {\em Nmap network scanning: The official Nmap project guide to network discovery and security scanning}.
\newblock Insecure, 2009.

\bibitem{Fscan}
2020.
\newblock \url{https://github.com/shadow1ng/fscan/blob/main/README_EN.md}.

\bibitem{longxiao2018webshell}
X.~Long, Y.~Fang, C.~Huang, and L.~Liu, ``Webshell research overview: The game between detection and evasion,'' {\em Cyberspace Security}, vol.~9, no.~1, pp.~62--68, 2018.

\bibitem{enoch2020harmer}
S.~Y. Enoch, Z.~Huang, C.~Y. Moon, D.~Lee, M.~K. Ahn, and D.~S. Kim, ``Harmer: Cyber-attacks automation and evaluation,'' {\em IEEE Access}, vol.~8, pp.~129397--129414, 2020.

\bibitem{schwartz2020pomdp+}
J.~Schwartz, H.~Kurniawati, and E.~El-Mahassni, ``Pomdp+ information-decay: Incorporating defender's behaviour in autonomous penetration testing,'' in {\em Proceedings of the International Conference on Automated Planning and Scheduling}, vol.~30, pp.~235--243, 2020.

\bibitem{becker2024evaluation}
N.~Becker, D.~Reti, E.~V. Ntagiou, M.~Wallum, and H.~D. Schotten, ``Evaluation of reinforcement learning for autonomous penetration testing using a3c, q-learning and dqn,'' {\em arXiv preprint arXiv:2407.15656}, 2024.

\bibitem{zhou2021autonomous}
S.~Zhou, J.~Liu, D.~Hou, X.~Zhong, and Y.~Zhang, ``Autonomous penetration testing based on improved deep q-network,'' {\em Applied Sciences}, vol.~11, no.~19, p.~8823, 2021.

\bibitem{sarraute2013automated}
C.~Sarraute, ``Automated attack planning,'' {\em arXiv preprint arXiv:1307.7808}, 2013.

\bibitem{xu2024autoattacker}
J.~Xu, J.~W. Stokes, G.~McDonald, X.~Bai, D.~Marshall, S.~Wang, A.~Swaminathan, and Z.~Li, ``Autoattacker: A large language model guided system to implement automatic cyber-attacks,'' {\em arXiv preprint arXiv:2403.01038}, 2024.

\bibitem{Cyberbattlesim}
W.~Blum, ``Gamifying machine learning for stronger security and ai models,'' {\em Microsoft Res., Redmond, WA, USA}, 2021.

\bibitem{sarraute2013penetration}
C.~Sarraute, O.~Buffet, and J.~Hoffmann, ``Penetration testing== pomdp solving?,'' {\em arXiv preprint arXiv:1306.4714}, 2013.

\bibitem{TenableNessus}
Tenable, ``Nessus vulnerability scanner: Network security solution.'' \url{https://www.tenable.com/products/nessus}, 2024.
\newblock Accessed: 2024-10-04.

\bibitem{MetasploitWebsite}
Rapid7, ``Metasploit - penetration testing software, pen testing security.'' \url{https://www.metasploit.com/}, 2024.
\newblock Accessed: 2024-10-04.

\bibitem{li2024knowledge}
Y.~Li, H.~Dai, and J.~Yan, ``Knowledge-informed auto-penetration testing based on reinforcement learning with reward machine,'' {\em arXiv preprint arXiv:2405.15908}, 2024.

\bibitem{zhang2022improved}
Y.~Zhang, J.~Liu, S.~Zhou, D.~Hou, X.~Zhong, and C.~Lu, ``Improved deep recurrent q-network of pomdps for automated penetration testing,'' {\em Applied Sciences}, vol.~12, no.~20, p.~10339, 2022.

\bibitem{guo2023automated}
X.~Guo, J.~Ren, J.~Zheng, J.~Liao, C.~Sun, H.~Zhu, T.~Song, S.~Wang, and W.~Wang, ``Automated penetration testing with fine-grained control through deep reinforcement learning,'' {\em Journal of Communications and Information Networks}, vol.~8, no.~3, pp.~212--220, 2023.

\bibitem{haeni1997firewall}
R.~E. Haeni, ``Firewall penetration testing,'' tech. rep., Technical report, The George Washington University Cyberspace Policy~…, 1997.

\bibitem{nmap}
``Nmap: the network mapper - free security scanner.'' \url{https://nmap.org/}.
\newblock Accessed: 2024-12-16.

\bibitem{mcdermott2001attack}
J.~P. McDermott, ``Attack net penetration testing,'' in {\em Proceedings of the 2000 workshop on New security paradigms}, pp.~15--21, 2001.

\bibitem{skaggs2002network}
B.~Skaggs, B.~Blackburn, G.~Manes, and S.~Shenoi, ``Network vulnerability analysis,'' in {\em The 2002 45th Midwest Symposium on Circuits and Systems, 2002. MWSCAS-2002.}, vol.~3, pp.~III--493, IEEE, 2002.

\bibitem{liu2005game}
P.~Liu, ``A game theoretic approach to cyber attack prediction,'' tech. rep., Pennsylvania State Univ., University Park, PA (United States), 2005.

\bibitem{kosuga2007sania}
Y.~Kosuga, K.~Kono, M.~Hanaoka, M.~Hishiyama, and Y.~Takahama, ``Sania: Syntactic and semantic analysis for automated testing against sql injection,'' in {\em Twenty-third annual computer security applications conference (ACSAC 2007)}, pp.~107--117, IEEE, 2007.

\bibitem{fonseca2007testing}
J.~Fonseca, M.~Vieira, and H.~Madeira, ``Testing and comparing web vulnerability scanning tools for sql injection and xss attacks,'' in {\em 13th Pacific Rim international symposium on dependable computing (PRDC 2007)}, pp.~365--372, IEEE, 2007.

\bibitem{shen2007strategies}
D.~Shen, G.~Chen, L.~Haynes, and E.~Blasch, ``Strategies comparison for game theoretic cyber situational awareness and impact assessment,'' in {\em 2007 10th International Conference on Information Fusion}, pp.~1--8, IEEE, 2007.

\bibitem{cone2007video}
B.~D. Cone, C.~E. Irvine, M.~F. Thompson, and T.~D. Nguyen, ``A video game for cyber security training and awareness,'' {\em computers \& security}, vol.~26, no.~1, pp.~63--72, 2007.

\bibitem{greenwald2009automated}
L.~Greenwald and R.~Shanley, ``Automated planning for remote penetration testing,'' in {\em MILCOM 2009-2009 IEEE Military Communications Conference}, pp.~1--7, IEEE, 2009.

\bibitem{sarraute2011algorithm}
C.~Sarraute, G.~Richarte, and J.~Luc{\'a}ngeli~Obes, ``An algorithm to find optimal attack paths in nondeterministic scenarios,'' in {\em Proceedings of the 4th ACM workshop on Security and artificial intelligence}, pp.~71--80, 2011.

\bibitem{van2013flipit}
M.~Van~Dijk, A.~Juels, A.~Oprea, and R.~L. Rivest, ``Flipit: The game of “stealthy takeover”,'' {\em Journal of Cryptology}, vol.~26, pp.~655--713, 2013.

\bibitem{chapman2014playing}
M.~Chapman, G.~Tyson, P.~McBurney, M.~Luck, and S.~Parsons, ``Playing hide-and-seek: an abstract game for cyber security,'' in {\em Proceedings of the 1st International Workshop on Agents and CyberSecurity}, pp.~1--8, 2014.

\bibitem{chapman2016cyber}
Chapman and M.~David, ``Cyber hide-and-seek,'' 2016.

\bibitem{ficco2017simulation}
M.~Ficco, M.~Chora{\'s}, and R.~Kozik, ``Simulation platform for cyber-security and vulnerability analysis of critical infrastructures,'' {\em Journal of computational science}, vol.~22, pp.~179--186, 2017.

\bibitem{casola2018towards}
V.~Casola, A.~De~Benedictis, M.~Rak, and U.~Villano, ``Towards automated penetration testing for cloud applications,'' in {\em 2018 IEEE 27th International Conference on Enabling Technologies: Infrastructure for Collaborative Enterprises (WETICE)}, pp.~24--29, IEEE, 2018.

\bibitem{ghanem2019reinforcement}
M.~C. Ghanem and T.~M. Chen, ``Reinforcement learning for efficient network penetration testing,'' {\em Information}, vol.~11, no.~1, p.~6, 2019.

\bibitem{paul2019strategic}
S.~Paul and Z.~Ni, ``A strategic analysis of attacker-defender repeated game in smart grid security,'' in {\em 2019 IEEE Power \& Energy Society Innovative Smart Grid Technologies Conference (ISGT)}, pp.~1--5, IEEE, 2019.

\bibitem{2019NIG}
T.~Y. Zhou, Y.~C. Zang, J.~H. Zhu, and Q.~X. Wang, ``Nig-ap: a new method for automated penetration testing,'' {\em Frontiers of Information Technology \& Electronic Engineering}, vol.~20, no.~9, pp.~1277--1288, 2019.

\bibitem{valea2020towards}
O.~Valea and C.~Opri{\c{s}}a, ``Towards pentesting automation using the metasploit framework,'' in {\em 2020 IEEE 16th International Conference on Intelligent Computer Communication and Processing (ICCP)}, pp.~171--178, IEEE, 2020.

\bibitem{bhattacharya2020automated}
A.~Bhattacharya, T.~Ramachandran, S.~Banik, C.~P. Dowling, and S.~D. Bopardikar, ``Automated adversary emulation for cyber-physical systems via reinforcement learning,'' in {\em 2020 IEEE International Conference on Intelligence and Security Informatics (ISI)}, pp.~1--6, IEEE, 2020.

\bibitem{nguyen2020multiple}
H.~V. Nguyen, H.~N. Nguyen, and T.~Uehara, ``Multiple level action embedding for penetration testing,'' in {\em Proceedings of the 4th International Conference on Future Networks and Distributed Systems}, pp.~1--9, 2020.

\bibitem{costa2020charles}
G.~Costa and A.~Valenza, ``Why charles can pen-test: an evolutionary approach to vulnerability testing,'' {\em arXiv preprint arXiv:2011.13213}, 2020.

\bibitem{bland2020machine}
J.~A. Bland, M.~D. Petty, T.~S. Whitaker, K.~P. Maxwell, and W.~A. Cantrell, ``Machine learning cyberattack and defense strategies,'' {\em Computers \& security}, vol.~92, p.~101738, 2020.

\bibitem{hu2020apu}
T.~Hu, T.~Zhou, Y.~Zang, Q.~Wang, and H.~Li, ``Apu-d* lite: Attack planning under uncertainty based on d* lite,'' {\em CMC-COMPUTERS MATERIALS \& CONTINUA}, vol.~65, no.~2, pp.~1795--1807, 2020.

\bibitem{qian2021ontology}
K.~Qian, D.~Zhang, P.~Zhang, Z.~Zhou, X.~Chen, and S.~Duan, ``Ontology and reinforcement learning based intelligent agent automatic penetration test,'' in {\em 2021 IEEE International Conference on Artificial Intelligence and Computer Applications (ICAICA)}, pp.~556--561, IEEE, 2021.

\bibitem{erdHodi2021simulating}
L.~Erd{\H{o}}di, {\AA}.~{\AA}. Sommervoll, and F.~M. Zennaro, ``Simulating sql injection vulnerability exploitation using q-learning reinforcement learning agents,'' {\em Journal of Information Security and Applications}, vol.~61, p.~102903, 2021.

\bibitem{ji2021optimal}
X.-P. Ji, W.~Tian, W.~Liu, and G.~Liu, ``Optimal attack strategy selection of an autonomous cyber-physical micro-grid based on attack-defense game model,'' {\em Journal of Ambient Intelligence and Humanized Computing}, vol.~12, pp.~8859--8866, 2021.

\bibitem{yamin2022use}
M.~M. Yamin and B.~Katt, ``Use of cyber attack and defense agents in cyber ranges: A case study,'' {\em Computers \& Security}, vol.~122, p.~102892, 2022.

\bibitem{confido2022reinforcing}
A.~Confido, E.~V. Ntagiou, and M.~Wallum, ``Reinforcing penetration testing using ai,'' in {\em 2022 IEEE Aerospace Conference (AERO)}, pp.~1--15, IEEE, 2022.

\bibitem{tran2022cascaded}
K.~Tran, M.~Standen, J.~Kim, D.~Bowman, T.~Richer, A.~Akella, and C.-T. Lin, ``Cascaded reinforcement learning agents for large action spaces in autonomous penetration testing,'' {\em Applied Sciences}, vol.~12, no.~21, p.~11265, 2022.

\bibitem{hance2022distributed}
J.~Hance, J.~Milbrath, N.~Ross, and J.~Straub, ``Distributed attack deployment capability for modern automated penetration testing,'' {\em Computers}, vol.~11, no.~3, p.~33, 2022.

\bibitem{faeroy2023automatic}
F.~L. F{\ae}r{\o}y, M.~M. Yamin, A.~Shukla, and B.~Katt, ``Automatic verification and execution of cyber attack on iot devices,'' {\em Sensors}, vol.~23, no.~2, p.~733, 2023.

\bibitem{li2023innes}
Q.~Li, M.~Hu, H.~Hao, M.~Zhang, and Y.~Li, ``Innes: An intelligent network penetration testing model based on deep reinforcement learning,'' {\em Applied Intelligence}, vol.~53, no.~22, pp.~27110--27127, 2023.

\bibitem{alshehri2024breachseek}
I.~Alshehri, A.~Alshehri, A.~Almalki, M.~Bamardouf, and A.~Akbar, ``Breachseek: A multi-agent automated penetration tester,'' {\em arXiv preprint arXiv:2409.03789}, 2024.

\bibitem{Zhenduo}
Z.~Wang, S.~Li, L.~Zhang, C.~Hu, and L.~Yan, ``A red team automated testing modeling and online planning method for post-penetration,'' {\em Computers \& Security}, p.~103945, 2024.

\bibitem{li2024dynpen}
Q.~Li, R.~Wang, D.~Li, F.~Shi, M.~Zhang, and A.~Chattopadhyay, ``Dynpen: Automated penetration testing in dynamic network scenarios using deep reinforcement learning,'' {\em IEEE Transactions on Information Forensics and Security}, 2024.

\bibitem{wang2024pentraformer}
Y.~Wang, S.~Liu, W.~Wang, C.~Zhu, C.~Fan, K.~Huang, and C.~Chen, ``Pentraformer: Learning agents for automated penetration testing via sequence modeling,'' in {\em 2024 IEEE International Conferences on Internet of Things (iThings) and IEEE Green Computing \& Communications (GreenCom) and IEEE Cyber, Physical \& Social Computing (CPSCom) and IEEE Smart Data (SmartData) and IEEE Congress on Cybermatics}, pp.~551--558, IEEE, 2024.

\bibitem{2024Graph}
Y.~Wang, S.~Liu, C.~Zhang, W.~Wang, J.~Jin, C.~Zhu, and C.~Zhou, ``Graph pre-training for reconnaissance perception in automated penetration testing,'' in {\em International Conference on Intelligent Computing}, 2024.

\bibitem{sarraute2012pomdps}
C.~Sarraute, O.~Buffet, and J.~Hoffmann, ``Pomdps make better hackers: Accounting for uncertainty in penetration testing,'' in {\em Proceedings of the AAAI Conference on Artificial Intelligence}, vol.~26, pp.~1816--1824, 2012.

\bibitem{al2008scalable}
M.~Al-Fares, A.~Loukissas, and A.~Vahdat, ``A scalable, commodity data center network architecture,'' {\em ACM SIGCOMM computer communication review}, vol.~38, no.~4, pp.~63--74, 2008.

\end{thebibliography}

 


\begin{IEEEbiography}[{\includegraphics[width=1in,height=1.25in,clip,keepaspectratio]{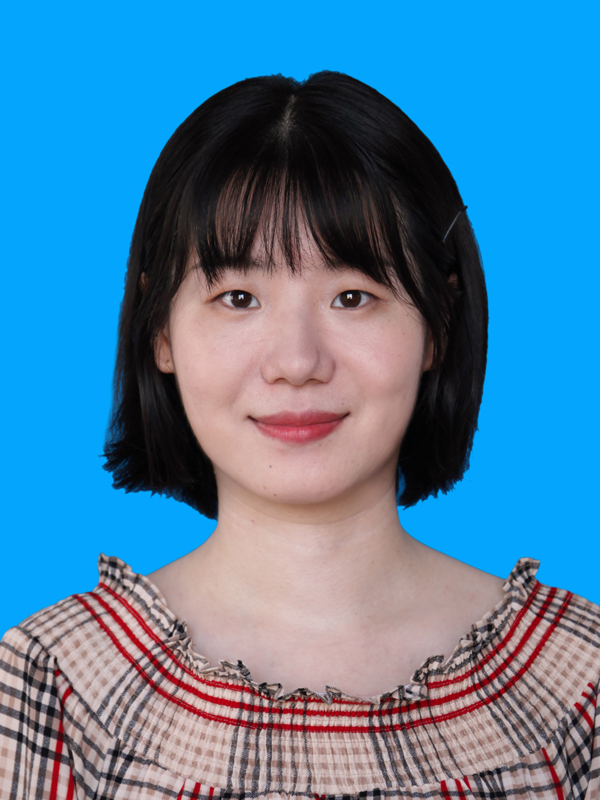}}]{Yunfei Wang}
received the B.S. degree in civil
engineering from the Hunan University, Changsha, China, in 2020. She is now pursuing the Ph.D degree at the National University of Defense Technology, Changsha, China. She is also a visiting scholar at Tsinghua University. Her research interests include auto penetration test, reinforcement learning and cyber-security.
\end{IEEEbiography}

\begin{IEEEbiography}[{\includegraphics[width=1in,height=1.25in,clip,keepaspectratio]{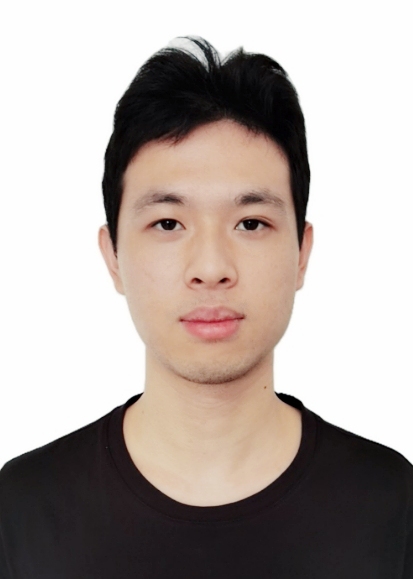}}]
{Shixuan Liu}
received his B.S. and Ph.D. degrees from the National University of Defense Technology, Changsha, China, in 2019 and 2024, respectively. He is also a visiting scholar in the Department of Computer Science and Technology at Tsinghua University, where he has spent two years. He has published over 10 papers in prestigious journals and conferences, including T-PAMI, T-KDE, T-CYB, and ICDM, focusing on knowledge reasoning and data mining.
\end{IEEEbiography}

\begin{IEEEbiography}[{\includegraphics[width=1in,height=1.25in,clip,keepaspectratio]{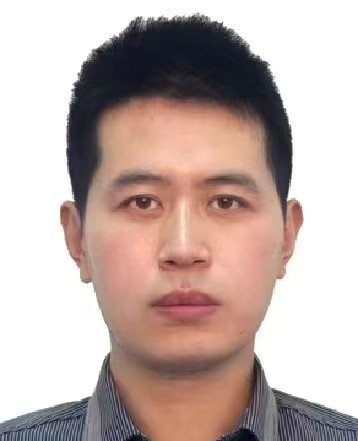}}]
{Wenhao Wang}
received the Ph.D. degrees from the National University of Defense Technology, Changsha, China. He is now a lecturer at the National University of Defense Technology. His research interests include auto penetration test, reinforcement learning and cyber-security.
\end{IEEEbiography}

\begin{IEEEbiography}[{\includegraphics[width=1in,height=1.25in,clip,keepaspectratio]{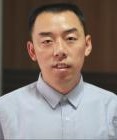}}]
{Changling Zhou}
received the B.S., M.S. and Ph.D. degrees from Peking University. He is a Professor with Peking University. His research interest include cyber-security, auto penetration test and LLM.
\end{IEEEbiography}

\begin{IEEEbiography}[{\includegraphics[width=1in,height=1.25in,clip,keepaspectratio]{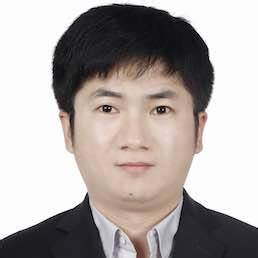}}]
{Chao Zhang} (Member, IEEE) received the B.S. and Ph.D. degrees from Peking University. He did Postdoctoral Research at UC Berkeley. He is an Associate Professor at Tsinghua University. His research interest lie in software and system security, including AI for security and security for AI.
\end{IEEEbiography}

\begin{IEEEbiography}[{\includegraphics[width=1in,height=1.25in,clip,keepaspectratio]{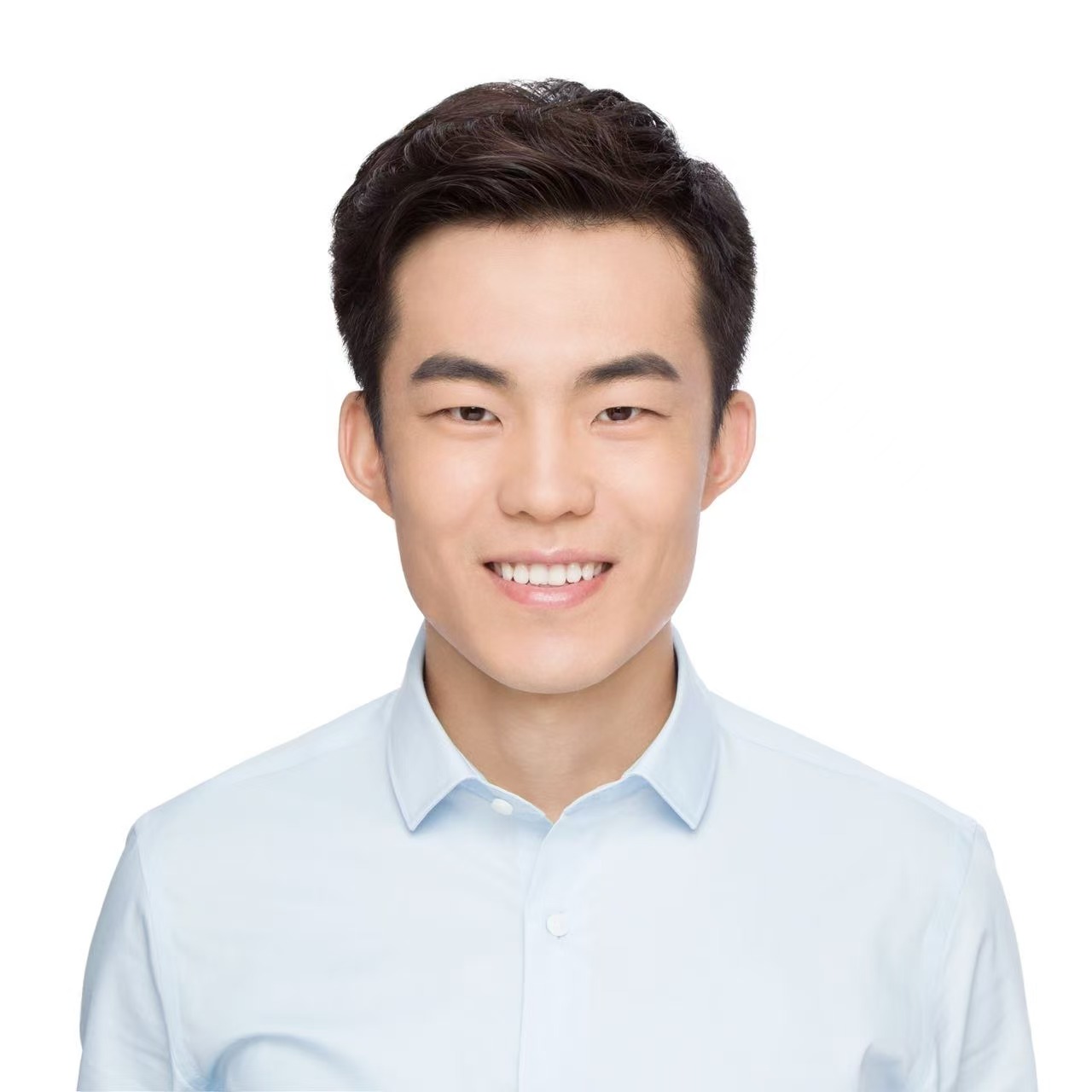}}]
{Jiandong Jin} received the M.S. and Ph.D. degrees from Peking University. He is an engineer in the Computer Center from Peking University.  His research interest include system security, including AI for security and security for AI.
\end{IEEEbiography}

\begin{IEEEbiography}[{\includegraphics[width=1in,height=1.25in,clip,keepaspectratio]{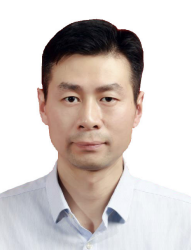}}]
{Cheng Zhu}
received the Ph.D. degree in management science and engineering from the National University of Defense Technology,
China, in 2005. He is currently a Professor with the National Key Laboratory of Information Systems Engineering, National University of Defense Technology. His current research interest is computer network data mining, command and control, and cyber security.
\end{IEEEbiography}

\vfill


\end{document}